\documentclass[lettersize,journal]{IEEEtran}
\usepackage{amsmath,amsfonts}
\usepackage{algorithmic}
\usepackage{array}
\usepackage[caption=false,font=normalsize,labelfont=sf,textfont=sf]{subfig}
\usepackage{textcomp}
\usepackage{stfloats}
\usepackage{url}
\usepackage{verbatim}
\usepackage{graphicx}
\def\BibTeX{{\rm B\kern-.05em{\sc i\kern-.025em b}\kern-.08em
    T\kern-.1667em\lower.7ex\hbox{E}\kern-.125emX}}
\usepackage{balance}

\usepackage{amssymb,graphicx,amsmath,algorithm,algorithmic,url} 
\usepackage{booktabs}
\usepackage{multirow}
\usepackage[shortlabels]{enumitem}
\usepackage{amssymb,xcolor,amsmath,bbm,multirow,bbold,mathtools,amssymb,graphics,graphicx,epstopdf,ntheorem,diagbox,enumitem,diagbox}
\usepackage{xr}
\usepackage{booktabs} 

\newtheorem{theorem}{Theorem}
\newtheorem{corollary}{Corollary}
\newtheorem{proposition}{Proposition}
\newtheorem*{proof}{Proof}
\theoremstyle{remark}

\DeclareMathOperator{\diag}{\rm diag}
\usepackage{color,comment}

\begin{document}
\title{Robust Subspace-Constrained Quadratic Models for Low-Dimensional Structure Learning}
\author{Zheng~Zhai and~Xiaohui~Li \thanks{ Zheng~Zhai is with Department of Statistics, Faculty of Arts and Sciences at Beijing Normal University, Zhuhai. Xiaohui~Li is with School of Mathematics and Information Sciences, Yantai University. }}

\markboth{Journal of \LaTeX\ Class Files,~Vol.~18, No.~9, September~2020}%
{How to Use the IEEEtran \LaTeX \ Templates}

\maketitle

\begin{abstract}
In this paper, we propose a robust subspace-constrained quadratic model (SCQM) for learning low-dimensional structure from high-dimensional data. Building upon the subspace-constrained quadratic matrix factorization (SQMF) framework, the proposed model accommodates a broad class of noise distributions, including generalized Gaussian and radial Laplace models. This generalization enables reliable performance under both heavy-tailed and light-tailed noise, thereby substantially enhancing robustness across diverse data regimes. To efficiently address the resulting nonconvex optimization problem, we develop a gradient-based algorithm equipped with a backtracking line-search strategy that ensures stable and efficient convergence. In addition, we present a sensitivity analysis of the $\ell_p^p$ and $\ell_2$ loss functions, elucidating their distinct behaviors under varying noise characteristics. Extensive numerical experiments corroborate the theoretical analysis and demonstrate that the proposed approach consistently outperforms existing methods in terms of robustness and reconstruction accuracy.
\end{abstract}


\section{Introduction}

\IEEEPARstart{L}{earning} low-dimensional structures from high-dimensional data remains a fundamental challenge in data analysis, machine learning, and signal processing. Traditional methods, including local linear fitting~\cite{lpca}, kernel density estimation (KDE)~\cite{genovese2014nonparametric}, linear matrix factorization, manifold fitting~\cite{fefferman2018fitting} and principal curves and surfaces~\cite{ozertem2011locally}, typically rely on the assumption that data reside in a linear subspace and are perturbed by Gaussian noise. While these assumptions enable the development of efficient algorithms, they are often unrealistic in real-world scenarios. In practice, data may lie on complex nonlinear manifolds and be affected by heavy-tailed noise or outliers. 

The assumption of flatness often leads to focusing on small, localized regions of the data. However, when narrowing the focus to such regions, there may be insufficient samples to apply these algorithms effectively, resulting in biased estimates. This limitation can cause a significant loss of information and degrade model performance. To overcome these challenges, it is imperative to develop models that can accommodate a broader range of data structures, capturing the intricate, nonlinear nature of real-world data while remaining robust to noise and outliers. Such models will be better equipped to provide more accurate, generalizable results and offer enhanced performance in practical applications.

To address this limitation, manifold learning and nonlinear factorization models have been proposed. Among them, quadratic matrix factorization (QMF)~\cite{zhai2024quadratic} and subspace-constrained quadratic matrix factorization (SQMF)~\cite{zhai2025subspace} have gained attention for their ability to incorporate second-order information and approximate curved manifolds. By augmenting linear subspace models~\cite{liu2010robust,zhang2017low} with quadratic terms, these methods can capture both tangent directions and curvature, improving reconstruction accuracy and interpretability. In particular, subspace-constrained quadratic matrix factorization (SQMF) provides a principled framework for learning low-dimensional tangent and normal spaces, as well as a quadratic mapping between them.

Existing quadratic factorization models often rely on squared Euclidean loss or the Frobenius norm, both of which assume Gaussian noise. While these assumptions offer computational simplicity, they make the models highly sensitive to outliers and misspecifications. To mitigate the impact of outliers, typical quadratic factorization models introduce a penalty term that penalizes the contribution of the quadratic term, aiming to prevent overfitting. However, this approach introduces its own challenges, such as the difficult task of selecting appropriate penalty parameters.

Another key reason why the Frobenius norm loss may be inappropriate is due to the mismatch between its assumptions and the characteristics of real-world noise distributions. In practical applications—such as image analysis, sensor data processing, and robust representation learning—data often contains noise that is heavy-tailed, sparse, or impulsive. These types of noise can severely degrade model performance and lead to biased estimates. The assumption of Gaussian noise, which is commonly used in the Frobenius norm loss, fails to capture the true nature of noise in many real-world scenarios. This discrepancy highlights the necessity for more flexible and robust loss functions within quadratic factorization models, ones that are specifically designed to account for the diverse and complex noise patterns encountered in practice. Such loss functions would improve the model's ability to handle these noise complexities, ultimately enhancing performance and robustness in real-world applications.

In response to these challenges, we propose a robust subspace-constrained quadratic learning framework that allows for a broader class of noise distributions, including generalized Gaussian and Laplace families. This flexibility enables the model to adapt to both light-tailed and heavy-tailed noise while maintaining the expressive power of quadratic manifold representations. Consequently, the proposed model offers enhanced robustness and accuracy in practical scenarios where Gaussian assumptions fail.

From an optimization perspective, we extend the classical Frobenius-norm formulation \(\|\cdot\|_{\rm F}^2\) to matrix factorization under alternative norms, such as the entrywise \(\ell_{1,1}\) norm and the mixed \(\ell_{2,1}\) norm. To solve the resulting optimization problem, we propose a gradient descent-based approach. Due to the quadratic structure of the mapping and the orthogonality constraints, the optimization remains nonconvex. To address this, we derive the Karush-Kuhn-Tucker (KKT) conditions and develop a Riemannian gradient descent algorithm on the Stiefel manifold, ensuring feasibility and convergence through orthogonality-preserving updates.

The main contributions of this work are as follows:
\begin{itemize}
    \item We propose a generalized subspace-constrained quadratic framework that extends existing SQMF models to a broad class of loss functions and derive explicit gradients for all variables, including latent coordinates.
    \item We provide a theoretical analysis, based on the implicit function theorem, that explains why the \(\ell_p^p\) loss and non-squared Euclidean norm improve robustness.
    \item We develop an efficient Riemannian gradient descent algorithm with orthogonality-preserving updates on the Stiefel manifold.
    \item Extensive numerical experiments demonstrate that the proposed method significantly improves robustness and reconstruction accuracy in the presence of outliers.
\end{itemize}

The structure of the paper is as follows. In Section~\ref{sec:model}, we introduce the subspace-constrained quadratic model, detailing its formulations and the associated identifiability property. We also provide a toy example featuring a circle in \(\mathbb{R}^2\) to demonstrate the model's effectiveness. Section~\ref{sec:gradients} focuses on deriving the gradients with respect to the free variables and orthogonal constraints, presenting the KKT conditions, and outlining the proposed optimization algorithm. In Section~\ref{sec:convexity}, we conduct a convexity analysis of the subproblems related to \(c\), \(\Theta\), and the projection with respect to \(\tau\). Section~\ref{sec:sensitivity} offers a sensitivity analysis for the \(\ell_p^p\) and \(\ell_2\) loss functions. Section~\ref{sec:experiments} presents numerical experiments to validate the theoretical results, and the paper concludes in Section~\ref{sec:conclusion} with a discussion of promising directions for future research.

\subsection{Related works}
Since our work leverages a subspace-constrained quadratic matrix factorization framework to learn latent manifold structures, it is closely related to a broad class of manifold fitting and denoising methods. These approaches aim to recover low-dimensional geometric structures embedded in high-dimensional observations, typically under noise and sampling irregularities. Representative techniques include local linear fitting methods such as local principal component analysis (LPCA)~\cite{lpca}, ridge-based approaches derived from kernel density estimation (KDE)~\cite{kde} and its logarithmic variant (LOG-KDE)~\cite{msf}, tangent-space aggregation methods such as manifold fitting (MFIT)~\cite{mfit}, and higher-order regression techniques including moving least squares (MLS)~\cite{mls}. For a comprehensive comparison, we focus on seven representative methods spanning these categories.

Ridge estimation constitutes one of the most influential paradigms for nonparametric manifold estimation. Rather than explicitly parameterizing the manifold, ridge-based methods define it implicitly as a set of points satisfying specific differential conditions of an estimated probability density function. In particular, KDE-based ridge estimation constructs a smooth density estimate from the observed data and identifies the manifold as a subset where the gradient aligns with the principal eigenspace of the Hessian, while the remaining curvature directions exhibit concavity. This formulation allows the manifold to be recovered directly from data geometry without requiring an explicit embedding or coordinate chart. In practice, the ridge set rarely admits a closed-form solution, and iterative procedures such as the subspace-constrained mean shift (SCMS)~\cite{ozertem2011locally, chen2025power} algorithm are commonly employed to trace the ridge structure. The LOG-KDE variant further modifies this framework by applying the ridge conditions to the logarithm of the density, leading to a different curvature characterization and tangent space estimation, often improving robustness in regions of varying density.

Beyond ridge-based techniques, several methods reconstruct manifolds by directly exploiting local tangent or normal space information. A prominent example is manifold fitting (MFIT), which estimates the manifold by enforcing consistency across locally estimated normal spaces. MFIT aggregates normal directions obtained from neighborhoods of nearby points using spatially adaptive weights and defines the manifold as the set of points where the weighted normal projections vanish. This approach provides a principled way to fuse local geometric information into a globally coherent manifold estimate and is particularly effective when normal directions can be reliably estimated from noisy data.

Higher-order manifold estimation methods aim to move beyond linear or first-order approximations by explicitly modeling local curvature. Among these, the moving least squares (MLS) framework is one of the most widely studied. MLS first estimates a local tangent space around each query point and then performs a weighted polynomial regression in this local coordinate system. The fitted polynomial captures higher-order geometric features, and the manifold estimate is obtained by projecting the query point onto the polynomial surface. Owing to its flexibility and strong approximation properties, MLS is well suited for smooth manifolds with nontrivial curvature, albeit at the cost of increased computational complexity and sensitivity to parameter choices.

Another closely related line of work extends local linear models by incorporating explicit geometric primitives. Spherical PCA (SPH), also known as spherelets, augments local PCA~\cite{lpca} by fitting low-dimensional spherical structures instead of affine subspaces. The method first projects data into a locally estimated affine subspace of dimension one higher than the intrinsic dimension and then fits a sphere within this space. By modeling curvature through spherical geometry, SPH provides a more accurate approximation than linear methods when the underlying manifold exhibits approximately constant curvature.

In contrast to the above approaches, which primarily rely on local geometric estimation and projection-based reconstruction, our method adopts the generalized quadratic approximation perspective with explicit subspace constraints. By integrating quadratic structure and subspace regularization, the proposed framework bridges manifold learning and data approximation, enabling robust recovery of latent manifolds while maintaining computational efficiency and scalability. This distinction allows our approach to complement existing local fitting and denoising methods, particularly in settings where global structure and latent representations are of primary interest.

\section{SCQM and its formulations}\label{sec:model}
In this section, we present the generalized SCQM model and explore various loss functions, highlighting their appropriate use cases and scenarios. We also provide a toy example to demonstrate the advantages of SCQM over the traditional linear local fitting model, illustrating how SCQM model can better capture complex patterns and improve performance in practical applications.
\subsection{Quadratic Fitting Model with Subspace Constraint}

We generalize classical quadratic matrix factorization beyond the Frobenius-norm objective by allowing a broad class of loss functions. This enables estimation in models of the form
\begin{equation}\label{eq:model_general}
x_i = f(\tau_i) + \epsilon_i, \qquad i=1,\dots,n,
\end{equation}
where the noise term $\epsilon_i$ is not restricted to be Gaussian. Instead, we allow $\epsilon_i$ to follow either heavy-tailed or light-tailed distributions. A representative example is the generalized Gaussian distribution with density
$p(\epsilon) \propto \exp \bigl(-\|\epsilon\|_p^{\,p}/\eta\bigr), 
\ x \in \mathbb{R}^D,$
which includes the Gaussian model when $p=2$ and the Laplace model when $p=1$. We also consider isotropic, rotation-invariant noise modeled by the multivariate radial Laplace distribution $p(\epsilon)\propto \exp(-\|\epsilon\|_2/\eta)$.

When the noise distribution is known, maximum likelihood estimation of the mapping $f(\cdot)$ reduces, up to additive constants, to the optimization problem
\begin{equation}\label{eq:mle_general}
\min_{f,\{\tau_i\}} \sum_{i=1}^n \ell\bigl(x_i - f(\tau_i)\bigr),
\end{equation}
where $\ell(\cdot)$ is the negative log-likelihood associated with the assumed noise model. Typical choices include $\|r\|_p^p$, $\|r\|_2$, and $\|r\|_2^2$, corresponding to generalized Gaussian, radial Laplace, and Gaussian noise models, respectively. This formulation establishes a direct link between the loss function and the statistical properties of the noise.

In what follows, we study the solution to~\eqref{eq:mle_general} by explicitly specifying both the loss function $\ell(\cdot)$ and the function class of $f$. In particular, we restrict $f$ to the class of quadratic functions equipped with explicit tangent- and normal-space basis representations. This choice offers two key advantages. First, it provides a clear geometric interpretation of the model parameters while retaining strong representation capability. Second, the linear model naturally arises as a special case of this framework by setting all quadratic terms to zero. As a result, this formulation allows us to seamlessly degenerate the quadratic model into its linear counterpart, thereby facilitating a direct and systematic investigation of the effectiveness and necessity of the quadratic terms.

\subsubsection{Quadratic Function Class}

To improve the quality of approximation over a broader domain, we restrict $f(\cdot)$ to the class of quadratic mappings for two main reasons. First, polynomial parameterizations offer significant practical advantages in optimization, as they lead to smooth objectives with well-structured gradients and Hessians. Second, the quadratic model is consistent with the manifold assumption, as it explicitly captures the interaction between the tangent space and the normal space through second-order terms. Specifically, we define the restricted class of quadratic mappings $f(\cdot)$ as follows:
\[
\begin{aligned}
\mathcal{F}
=
\bigl\{
f:\mathbb{R}^d \to \mathbb{R}^D
\;\big|\;
f(\tau) = c + U\tau + V\,\mathcal{A}(\tau,\tau), \\
\qquad\qquad
U^\top U = I_d,\;
V^\top V = I_s,\;
U^\top V = 0
\bigr\}.
\end{aligned}
\]
Here, $c$ denotes the shift parameter. The columns of $U$ form an orthonormal basis of the tangent space, while the columns of $V$ span a subspace orthogonal to that of $U$. The vector $\mathcal{A}(\tau,\tau) \in \mathbb{R}^s$ represents the action of a third-order tensor $\mathcal{A}$ on $(\tau,\tau)$, defined element-wise by
\[
\bigl\{\mathcal{A}(\tau,\tau)\bigr\}_k
=
\sum_{i,j=1}^d \mathcal{A}_{k,i,j}\,\tau_i\tau_j,
\qquad k=1,\dots,s.
\]
Since the rank-one matrix $\tau\tau^\top$ is symmetric, we may assume without loss of generality that each slice $\mathcal{A}_k \in \mathbb{R}^{d\times d}$ is symmetric. By concatenating $U$ and $V$ into a single matrix $Q=[U,V]\in\mathbb{R}^{D\times(d+s)}$, the orthogonality constraints can be written compactly as $Q^\top Q = I_{d+s}$. This yields the equivalent representation
\begin{equation}\label{eq:f_compact}
f(\tau)
=
c
+
Q
\begin{bmatrix}
\tau \\
\mathcal{A}(\tau,\tau)
\end{bmatrix},
\qquad
Q^\top Q = I_{d+s}.
\end{equation}
Exploiting the symmetry of $\mathcal{A}$, there exists a matrix $\Theta$ such that
$\mathcal{A}(\tau,\tau)
=
\Theta^\top \mathrm{vech}(\tau\tau^\top),$
where $\mathrm{vech}(\cdot)$ stacks the upper-triangular entries of a symmetric matrix into a vector. Consequently, $\mathcal{F}$ admits the equivalent formulation
\begin{equation}\label{eq:f_theta}
f(\tau)
=
c
+
Q
\begin{bmatrix}
\tau \\
\Theta^\top \mathrm{vech}(\tau\tau^\top)
\end{bmatrix},
\qquad
Q^\top Q = I_{d+s}.
\end{equation}
Our learning problem is:
\begin{equation}\label{eq:loss_general}
(\widehat{f},\  \{\widehat{\tau_i}\}) = \arg\min_{f \in \mathcal{F},\, \{\tau_i\}}
\;
\sum_{i=1}^n
\ell \bigl(x_i - f(\tau_i)\bigr).
\end{equation}
This model can be viewed as a generalized formulation of the subspace-constrained quadratic factorization problem studied in~\cite{zhai2025subspace}, in which the Frobenius norm is replaced by a column-wise loss defined through a specialized norm.

From Eqn.~\eqref{eq:loss_general}, we observe that the model simultaneously learns both the local geometry, represented by the parameters \( c \), \( Q \), and \( \Theta \) in the function \( f \), as well as the coordinates \( \{\tau_i\}_{i=1}^n \) derived from the high-dimensional input data \( \{x_i\}_{i=1}^n \). Once we obtain \( \widehat{f} \), we can use it as a model to refine the new noisy data \( y \) by projecting it onto \( \widehat{f} \), which is achieved by solving the following optimization problem:
\begin{equation}\label{projection}
\widehat{y} = \arg \min_{x = \widehat{f}(\tau)} \ell(x - y).
\end{equation}

From an algorithmic perspective, due to the nonlinearity of both the quadratic mapping and the general loss function, closed-form solutions are generally not available for the optimization problem in Eqn.~\eqref{eq:loss_general} and Eqn.~\eqref{projection}. In the following sections, we first address the identifiability issue and discuss principled criteria for selecting an appropriate loss function. We then propose an efficient gradient-based algorithm to solve the resulting optimization problem numerically.

\subsection{Identifiability}

The optimization problem in Eqn.~\eqref{eq:loss_general} is not identifiable due to the inherent invariances in the latent representation. Specifically, the model exhibits invariance under orthogonal transformations of the latent variables.

Let \( R \in O(d) \) be any orthogonal matrix, and define the transformed latent variable \( \eta = R\tau \). Since
\[
\eta \eta^\top = R(\tau\tau^\top)R^\top,
\]
there exists a matrix \( S(R) \) such that $\mathrm{vech}(\eta\eta^\top) = S(R)\,\mathrm{vech}(\tau\tau^\top).$
Consequently, by defining $U_R = U R^T,  \Theta_R = S(R)^{-\top}\Theta,$ we obtain
\[
\Theta^\top \mathrm{vech}(\tau\tau^\top) = \Theta_R^\top \mathrm{vech}(\eta\eta^\top),\qquad U\tau = U_R \eta.
\]
This implies that different parameter tuples \( (c, U, V, \Theta, \tau) \) and \( (c, U_R, V, \Theta_R, \eta) \), related by orthogonal transformations \( \eta = R\tau \), induce the same input-output mapping \( f(\tau) = f_R(\eta) \), where \( f_R(\cdot) \) denotes the function parameterized by \( c, U_R, V, \Theta_R \).

Therefore, the parameters of the quadratic factorization model are identifiable only up to equivalence classes induced by orthogonal transformations in the latent space. Although a unique parameterization cannot be recovered, the learned mapping \( f(\cdot) \) and the associated quadratic manifold are uniquely determined within these equivalence classes.

\subsection{Loss functions}
In this section, we investigate how to choose an appropriate loss function for the quadratic factorization model. The central principle is that the loss function should be aligned with the statistical distribution of the noise in the data or observations. Specifically, when the noise exhibits a long-tailed (heavy-tailed) distribution, an $\ell_p^p$ loss with $p<2$ provides a more robust and suitable choice. In contrast, when the noise follows a short-tailed distribution, the Gaussian assumption becomes appropriate, and the corresponding $\ell_2^2$ loss is a natural and effective option. This distribution-aware selection enables the model to better capture the underlying data characteristics and achieve improved robustness and accuracy.

\begin{table*}[t]
\centering
\caption{Common loss functions, together with their gradients and Hessians with respect to $x$, and the corresponding noise distributions under a maximum likelihood interpretation.}
\resizebox{\linewidth}{!}{
\renewcommand{\arraystretch}{1.45}
\small
\begin{tabular}{c||c|c|c|c}
\hline\hline
\textbf{Loss function} 
& $\boldsymbol{\ell(x-y)}$ 
& $\boldsymbol{\nabla_x \ell(x-y)}$ 
& $\boldsymbol{\nabla_x^2 \ell(x-y)}$
& \textbf{Suitable Distribution} \\ 
\hline
$\ell_1$ (Manhattan) 
& $\displaystyle \sum_{k=1}^d |x_k-y_k|$  
& $\mathrm{sign}(x-y)$ \textit{(subgradient)} 
& $0$ a.e. \textit{(undefined at $x=y$)}
& Independent Laplace \\ 
\hline
$\ell_2$ (Euclidean) 
& $\displaystyle \|x-y\|_2$  
& $\displaystyle \frac{x-y}{\|x-y\|_2},\quad x \neq y$
& $\displaystyle 
\frac{1}{\|x-y\|_2}
\!\left(
I-\frac{(x-y)(x-y)^{\!T}}{\|x-y\|_2^2}
\right)$
& Multivariate (isotropic) Laplace \\ 
\hline
$\ell_2^2$ (Squared Euclidean) 
& $\displaystyle \|x-y\|_2^2$  
& $2(x-y)$ 
& $2I$
& Multivariate Gaussian \\ 
\hline
$\ell_p^p$ ($p \ge 1$) 
& $\displaystyle \sum_{k=1}^d |x_k-y_k|^p$   
& $\displaystyle p\,|x-y|^{p-1} \odot \mathrm{sign}(x-y)$
& $\displaystyle
p(p-1)\,
\diag\!\bigl(|x-y|^{p-2}\bigr)$
& Generalized Gaussian \\ 
\hline
Mahalanobis (squared) 
& $\displaystyle (x-y)^T M (x-y)$ 
& $\displaystyle 2M(x-y)$
& $2M$
& Multivariate Gaussian \\ 
\hline
Huber distance
& $\displaystyle 
\begin{cases}
\frac{1}{2}\|x-y\|_2^2, & \|x-y\|_2 \le \delta, \\[4pt]
\delta\|x-y\|_2 - \frac{1}{2}\delta^2, & \|x-y\|_2 > \delta
\end{cases}$ 
& $\displaystyle
\begin{cases}
x-y, & \|x-y\|_2 \le \delta, \\[4pt]
\delta\,\dfrac{x-y}{\|x-y\|_2}, & \|x-y\|_2 > \delta
\end{cases}$ 
& $\displaystyle
\begin{cases}
I, & \|x-y\|_2 \le \delta, \\[4pt]
\delta\,\nabla^2\|x-y\|_2, & \|x-y\|_2 > \delta
\end{cases}$
& Gaussian--Laplace hybrid \\ 
\hline\hline
\end{tabular}}
\label{tab:loss_gradients}
\end{table*}

\subsection{Criteria for Choosing the Loss Function}
We discuss the criteria for loss-function selection under three different scenarios. First, we consider the simplest setting, in which the noise distribution ${\epsilon_i}$ is known \emph{a priori}. Second, we study the case where the contaminated noise samples ${\epsilon_i}$ are directly observable. Third, we address the most realistic scenario, in which the noise is implicitly embedded in the observations and cannot be isolated from the underlying signal.

\paragraph{Known noise distribution}
In the first scenario, we select the loss function $\ell(\cdot)$ according to a principled, likelihood-based criterion under known noise assumptions. By the law of large numbers, as the sample size increases, the empirical distribution of the noise converges to its true underlying distribution. Consequently, a statistically sound and natural choice of the loss function is the negative log-likelihood induced by the assumed noise model. This choice leads to statistically consistent estimators and, with high probability, ensures that the optimization objective faithfully reflects the intrinsic characteristics of the noise. Guided by this principle, we consider several commonly adopted noise models together with their corresponding loss functions in the following analysis.

When the noise is isotropic Gaussian, i.e., $\epsilon_i \sim \mathcal{N}(0,\sigma^2 I)$, the negative log-likelihood is proportional to the squared Euclidean norm. Accordingly, we adopt the loss $\ell(r)=\|r\|_2^2$, which recovers the classical least-squares formulation. If the noise instead follows a multivariate isotropic (radial) Laplace distribution with density $p(\epsilon)\propto \exp(-\lambda\|\epsilon\|_2)$, the corresponding negative log-likelihood leads to the Euclidean norm loss $\ell(r)=\|r\|_2$. When the noise components are independently distributed according to univariate Laplace laws, the negative log-likelihood becomes proportional to the $\ell_1$ norm, and we employ $\ell(r)=\|r\|_1$, which is well known for its robustness to sparse, large-magnitude outliers.

More generally, heavy-tailed or light-tailed noise can be modeled by a generalized Gaussian distribution with density $p(\epsilon)\propto \exp(-|\epsilon|_p^p)$ for $p\ge1$, motivating the loss $\ell(r)=\|r\|_p^p$. Smaller values of $p$ yield increased robustness to outliers, while larger values recover the Gaussian setting. When the noise covariance is anisotropic, i.e., $\epsilon_i\sim\mathcal{N}(0,\Sigma)$ with $\Sigma\succ0$, the appropriate choice is the squared Mahalanobis loss $\ell(r)=r^\top\Sigma^{-1}r$, which accounts for correlations and heterogeneous scaling across dimensions. Finally, in the presence of mixed noise models or outlier contamination, the Huber loss is frequently adopted, as it interpolates smoothly between the $\ell_2$ and $\ell_1$ norms, combining robustness with local smoothness.

Table~\ref{tab:loss_gradients} summarizes several commonly used loss functions together
with their associated noise models from a maximum-likelihood perspective. Overall, the
choice of the loss function $\ell(\cdot)$ plays a central role in balancing statistical
efficiency and robustness, and is therefore critical to the performance of the proposed
quadratic factorization framework.

\paragraph{Observable noise samples}
In the second scenario, where the noise samples $\{\epsilon_i\}_{i=1}^n$ are explicitly observable, the optimal shape parameter $p$ in the $\ell_p^p$ loss can be estimated via maximum likelihood estimation (MLE)~\cite{roenko2014estimation}. Since no closed-form solution is available, we suggest to employ a numerical optimization procedure based on the profile likelihood. In particular, a one-dimensional search strategy, such as grid search or bisection, is adopted, following standard practice in the literature~\cite{pascal2013parameter,do2002wavelet}.

\paragraph{Unobservable noise embedded in observations.}
In the third and most practical scenario, the noise is intrinsically mixed with the unknown underlying signal, rendering direct maximum likelihood estimation of the noise distribution infeasible. Nevertheless, meaningful and practically effective criteria can still be employed for loss-function selection. In this case, we propose the following strategies:

\begin{enumerate}
\item \textbf{Control of heavy-tailed noise}
Employing the $\ell_p^p$ loss with a relatively small value of $p$ mitigates the influence of heavy-tailed noise. This choice reduces sensitivity to samples that lie far from the estimated curve, thereby enhancing robustness to large deviations.

\item \textbf{Euclidean-distance-based projection.}
Both the $\ell_2^2$ and $\ell_2$ losses yield projections that minimize the Euclidean distance between data points and the estimated model, aligning with the classical notion of projection. In particular, the $\ell_2^2$ loss corresponds to the maximum likelihood estimator under Gaussian noise, which is widely adopted in practice.

\item \textbf{A conservative choice.}
The $\ell_2$ loss serves as a conservative and stable choice, as it exhibits reliable performance across a wide range of noise conditions, including light-tailed and moderately heavy-tailed distributions. Moreover, when computing the optimal local coordinates associated with the fitted low-dimensional surface (or curve in $\mathbb{R}^2$), the $\ell_2$ and $\ell_2^2$ losses are equivalent in the sense that they induce the same minimizers. Consequently, both losses admit the standard distance-based interpretation of orthogonal projection, which is consistent with classical geometric formulations.
\end{enumerate}

    

\begin{figure}[t]
    \centering
\includegraphics[width=\linewidth]{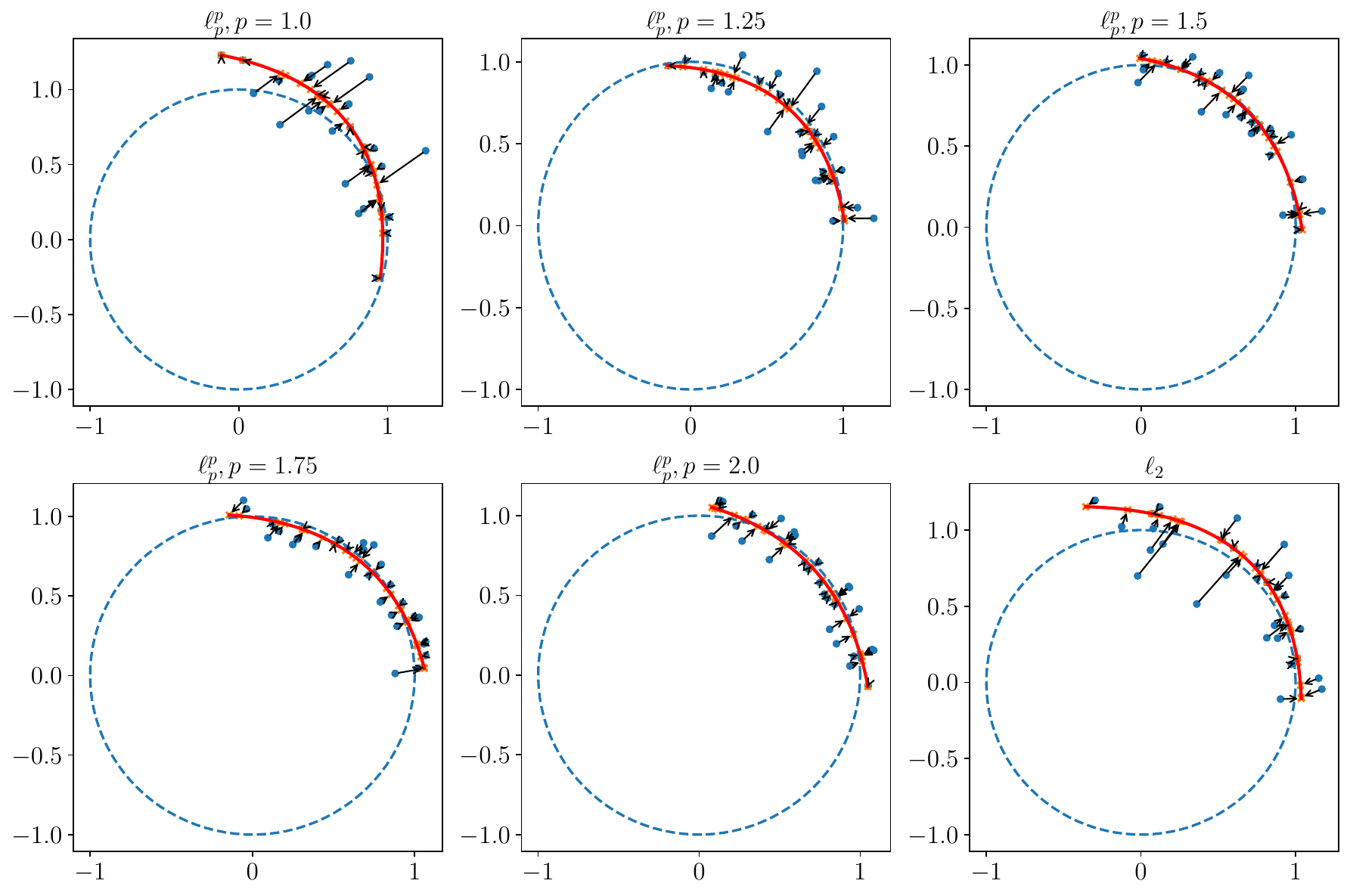}
    \caption{Illustration of the fitted curves and projection points obtained using $\ell_p^p$ losses with different values of $p$.}
    \label{fig:robustness2}
\end{figure}
\subsection{A toy example in ${\mathbb R}^2$} \label{toy}
In this experiment, we demonstrate that when the loss function is chosen to match the underlying noise distribution, all models achieve strong performance. We present a synthetic example demonstrating that the choice of loss function should be consistent with the underlying noise distribution. Specifically, we generate data according to
\begin{equation}
x_i = [\cos(t_i), \sin(t_i)]^T + \epsilon_i,
\qquad
\epsilon_i \sim C_{p,d}\exp\bigl(-\|\epsilon_i\|_p^p\bigr),
\end{equation}
where $C_{p,d}$ denotes the normalizing constant, $\{t_i\}$ consists of equally spaced points in the interval $[0,4]$, and $\epsilon_i$ follows a multivariate generalized Gaussian distribution with shape parameter $p$.
Also, to assess the performance of the $\ell_2$ loss under a matched noise model, we additionally consider a setting in which the noise follows a multivariate isotropic Laplace distribution, namely,
$
\epsilon_i \sim C'_d \exp\bigl(-\|\epsilon_i\|_2\bigr),
$
which corresponds to the negative log-likelihood associated with the Euclidean norm.

\begin{figure}[t]
    \centering
\includegraphics[width=\linewidth]{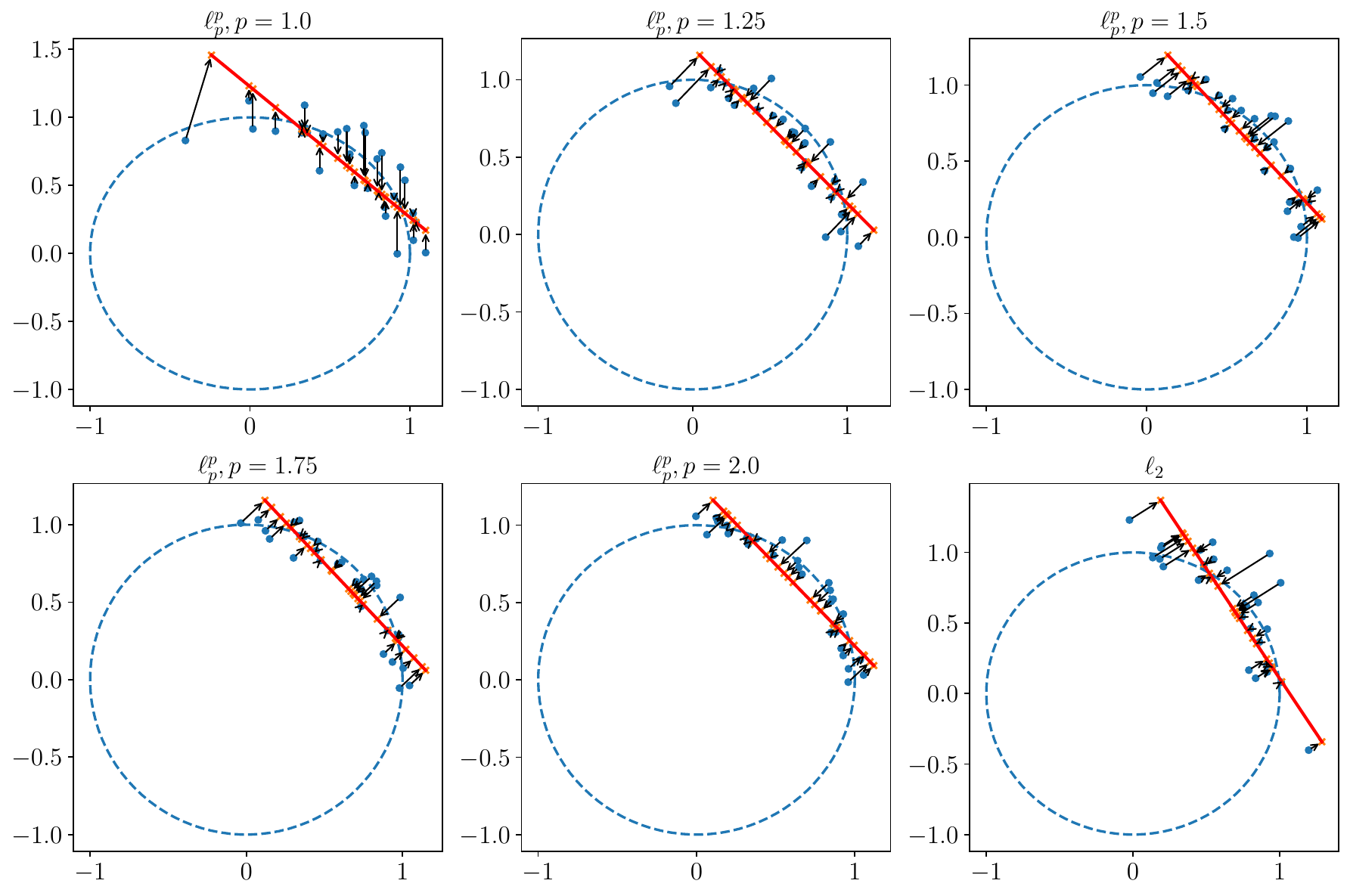}
    \caption{Illustration of the fitted curves and projection points obtained using $\ell_p^p$ losses with different values of $p$ when restricting the function class to be linear (${\cal A} =\bf 0$).}
    \label{fig:robustness2_3d}
\end{figure} 

To this end, we describe how to generate synthetic noise samples whose density is proportional to $\exp(-\|\epsilon\|_p^p)$. We focus on the multivariate generalized Gaussian distribution~\cite{kotz2012laplace}, whose probability density function takes exactly this form. Owing to its separability, the distribution factorizes into independent generalized Gaussian marginals along each coordinate. Consequently, a random vector in $\mathbb{R}^d$ can be generated by independently sampling each coordinate from the one-dimensional density
$f(t) \propto \exp \left(-|t|^p\right).$
In practice, such samples can be efficiently constructed using a simple transformation: draw $u \sim \mathrm{Gamma}(1/p, 1)$ and an independent Rademacher random variable $s \in \{-1,1\}$ with equal probability, and set $\epsilon = s\,u^{1/p}$. Repeating this procedure independently for each coordinate yields a random vector whose joint distribution follows the multivariate generalized Gaussian distribution.

In Fig.~\ref{fig:robustness2}, the fitted model $f$ is shown by the red curve, while the blue points denote the noisy observations. The arrows indicate the orthogonal projections of the noisy data points onto the underlying low-dimensional quadratic manifold defined by $f$. The noisy data are generated under different values of the shape parameter $p \in \{1.0, 1.25, 1.5, 1.75, 2.0\}$, and the corresponding $\|\cdot\|_p^p$ loss is employed in each case.

We make the following observations. First, the quadratic subspace-constrained model exhibits substantially stronger fitting capability than its linear counterpart, achieving tighter approximation over a broader domain. Second, the underlying low-dimensional structure can be successfully recovered by the proposed subspace-constrained quadratic matrix factorization model, provided that the loss function is properly matched to the noise distribution. In particular, for heavy-tailed noise, we recommend using the $\ell_p^p$ loss with smaller values of $p$ to enhance robustness, whereas for light-tailed noise, larger values of $p$, such as the squared Euclidean loss, are more suitable and yield higher statistical efficiency.



\section{Gradients and KKT condition} \label{sec:gradients}
In this section, we derive closed-form expressions for $\nabla_{\tau} F$, $\nabla_{\Theta} F, \nabla_{c} F$, and $\nabla_{Q} F$. The gradient $\nabla_{\tau} F$ is nontrivial to compute due to the involvement of the element-extraction operator $\mathrm{vech}(\cdot)$.
\subsection{Gradient with respect to $\{\tau_k\}_{k=1}^n$}
Here, we derive the gradient of the loss function with respect to the latent variable $\tau_k$. Owing to the separability of the loss across data samples, the objective function can be decomposed into independent terms, each depending only on a single latent representation. Consequently, the gradient with respect to $\tau_k$ can be computed by differentiating only the $k$-th loss term.
\[
\nabla_{\tau_k} F = \nabla_{\tau_k}\sum_{i=1}^n \ell(x_i-f(\tau_i)) = \nabla_{\tau_k}\ell(x_k-f(\tau_k)).
\]
Due to the nonlinear dependence of $f(\tau_k)$ on $\tau_k$ through 
$\mathrm{vech}(\tau_k \tau_k^T)$, a direct computation of 
$\nabla_{\tau_k} f(\tau_k)$ is nontrivial. 
Observing that, for a small perturbation $\delta \in \mathbb{R}^d$, the function $f(\tau_k+\delta)$ admits the following first-order expansion:
\begin{equation}\label{1st_order}
f(\tau_k+\delta)
=
f(\tau_k)
+
\langle \nabla_{\tau_k} f(\tau_k), \delta \rangle
+
\mathcal{O}(\|\delta\|_2^2).
\end{equation}
Based on ~\eqref{1st_order}, we derive the gradient by analyzing the difference 
$f(\tau_k+\delta)-f(\tau_k)$ by
\[
\begin{aligned}
&f(\tau_k+\delta) - f(\tau_k)\\
=&U\delta +V\Theta^T \mathrm{vech}((\tau_k+\delta)(\tau_k+\delta)^T)+V\Theta^T \mathrm{vech}(\tau_k\tau_k^T)\\
=&U\delta +V\Theta^T \mathrm{vech}(\tau_k\delta^T+\delta\tau_k^T+\delta\delta^T).
\end{aligned}
\]
Define the operators $T_\tau(\delta) = \mathrm{vech}(\tau \delta^T)$ and $T’_\tau(\delta) = \mathrm{vech}(\delta \tau^T)$. Observe that both $\mathrm{vech}(\tau \delta^T)$ and $\mathrm{vech}(\delta \tau^T)$ induce linear mappings with respect to the perturbation $\delta$. In particular, for any $\delta_1, \delta_2 \in \mathbb{R}^d$ and any scalar $\kappa \in \mathbb{R}$, these operators satisfy the additivity and homogeneity properties,
\[
T_{\tau_k}(\kappa\delta)
=
\mathrm{vech}\!\bigl(\tau_k(\kappa\delta)^T\bigr)
=
\kappa\mathrm{vech}(\tau_k\delta^T) = \kappa T_{\tau_k}(\delta).
\]
\[
\begin{aligned}
T_{\tau_k}(\delta_1+\delta_2) = &\mathrm{vech}\!\bigl(\tau_k(\delta_1+\delta_2)^T\bigr)\\
= &
\mathrm{vech}(\tau_k\delta_1^T)+\mathrm{vech}(\tau_k\delta_2^T)\\
= &T_{\tau_k}(\delta_1)+T_{\tau_k}(\delta_2).
\end{aligned}
\]

Therefore, both $T_{\tau_k}$ and $T'_{\tau_k}$ are linear mappings from $\mathbb{R}^d$ to $\mathbb{R}^{\frac{d(d+1)}{2}}$. 
By the matrix representation theorem for linear transformations~\cite{strang2022introduction}, any linear operator admits a matrix representation once the input and output bases are fixed. In the following, we adopt the standard bases and derive the explicit matrix form of the corresponding linear transformation.

\begin{corollary}[Matrix representation theorem] \label{cor1}
Let $T: \mathbb{R}^n \to \mathbb{R}^m$ be a linear transformation. Then there exists a unique matrix $A \in \mathbb{R}^{m\times n}$ such that
\[
T(x) = Ax, \quad \forall x \in \mathbb{R}^n.
\]
\end{corollary}

In order to find the corresponding transformation matrix. We take the standard basis of  $\{e_1,e_2,...,e_d\} \in \mathbb{R}^d$, Then, any $\delta \in  \mathbb{R}^d$ can be written as
\[
\begin{aligned}
T_{\tau_k}(\delta) =& T_{\tau_k}(\sum_{i=1}^d \langle\delta, e_i \rangle e_i) = \sum_{i=1}^d \langle\delta, e_i \rangle T_{\tau_k}(e_i) \\
=& [T_{\tau_k}(e_1), T_{\tau_k}(e_2),...,T_{\tau_k}(e_d)] \delta.
\end{aligned}
\]
Therefore, by defining $M_{\tau_k}=[T_{\tau_k}(e_1), T_{\tau_k}(e_2),...,T_{\tau_k}(e_d)]$ and $N_{\tau_k}=[T'_{\tau_k}(e_1), T'_{\tau_k}(e_2),...,T'_{\tau_k}(e_d)]$, we have:
\[
f(\tau_k+\delta) - f(\tau_k)=U\delta+V\Theta^T(M_{\tau_k}+N_{\tau_k})\delta + V\Theta^T \mathrm{vech}(\delta\delta^T).
\]
By the definitions of the linear operators $T(\cdot)$ and $T'(\cdot)$, the explicit forms of $T_{\tau_k}(e_i)$ and $T'_{\tau_k}(e_i)$ can be readily derived for all $i=1,\ldots,d$. Neglecting the higher-order term $V\Theta^T \mathrm{vech}(\delta\delta^T)$ in the first-order expansion, we obtain 
\[
\nabla_\tau f(\tau) = U+V\Theta^T(M_{\tau}+N_{\tau}).
\] Therefore, the gradient $\nabla_\tau F$ is:
\begin{equation}\label{grad_tau}
\begin{aligned}
\nabla_{\tau_k} F(c,Q,\Theta,\Phi) =& \nabla_{\tau_k}  \ell(x_k-f(\tau_k)) \\
= &{\nabla ^T_{\tau_k}f(\tau_k)}\nabla_y \ell(y-x_k)|_{y=f(\tau_k)}.
\end{aligned}
\end{equation}

Note that the Jacobian matrix $\nabla_{\tau_k} f(\tau_k)$ is defined with respect to the $k$-th latent representation $\tau_k$ corresponding to $x_k$. In the following, we present the general forms of the matrices $M_{\tau}$ and $N_{\tau}$, which characterize the linear mappings induced by the quadratic term. By stacking the vectors $T_\tau(e_j), j=1,\ldots,d$, as the columns of $M_\tau\in {\mathbb R}^{\{(d^2+d)/2\}\times d}$, and $T'_\tau(e_j), j=1,\ldots,d$, as the columns of $N_\tau\in {\mathbb R}^{\{(d^2+d)/2\}\times d}$, we obtain explicit matrix representations of the two linear operators. Consequently, for any $\tau \in \mathbb{R}^d$, the transformation matrices $M_\tau$ and $N_\tau$ admit the following structured forms:

{\small 
\[
M_{\tau} = 
\left( 
\begin{array}{ccccccc}
\tau_{[1]}&0 & 0& 0&0 & 0&0\\
0 & \tau_{[1]}& 0&0 & 0&0 & 0 \\
& & ...&... & & & \\
0&0 & 0&0 &0 &0 &\tau_{[1]}\\
0&\tau_{[2]}&0 & 0& 0&0 & 0\\
0 & 0& \tau_{[2]}&0 & 0&0 & 0 \\
& & ...& ...& & & \\
0&0 & 0&0 &0 &0 &\tau_{[2]}\\
& & ...& ...& & & \\
0&0 & 0&0 &0 &0 &\tau_{[d]}
\end{array} 
\right),
\]
\[
N_{\tau} = 
\left( 
\begin{array}{ccccccc}
\tau_{[1]}&0 & 0& 0&0 & 0&0\\
\tau_{[2]} & 0& 0&0 & 0&0 & 0 \\
& & ...&... & & & \\
\tau_{[d]}&0 & 0&0 &0 &0 &0\\
0&\tau_{[2]}&0 & 0& 0&0 & 0\\
0 & \tau_{[3]}& 0&0 & 0&0 & 0 \\
& & ...& ...& & & \\
0&\tau_{[d]} & 0&0 &0 &0 &0\\
& & ...& ...& & & \\
0&0 & 0&0 &0 &0 &\tau_{[d]}
\end{array} 
\right).
\]}
Noting that both $M_{\tau}$ and $N_{\tau}$ are highly sparse matrices, each containing exactly one nonzero entry per row, we can express them in the following compact form:
{\small \[
 \begin{gathered}
 N_{\tau}
=
[
\tau_{[1]} e_1,
\tau_{[2]} e_1,
\cdots,
\tau_{[d]} e_1,
\tau_{[2]} e_2,
\cdots,
\tau_{[d]} e_2,
\cdots,
\tau_{d} e_d
]^T, \\
M_{\tau}
=
[
\tau_{[1]} e_1,
\tau_{[1]} e_2,
\cdots,
\tau_{[1]} e_d,
\tau_{[2]} e_2,
\cdots,
\tau_{[2]} e_d,
\cdots,
\tau_{[d]} e_d]^T.
\end{gathered}
\]}
Here, $\tau_{[k]} \in \mathbb{R}$ denotes the $k$-th scalar component of $\tau \in \mathbb{R}^d$, which is distinguished from $\tau_k \in \mathbb{R}^d$, the $k$-th latent vector associated with the data point $x_k \in \mathbb{R}^D$. The vector $e_k$ denotes the $k$-th canonical basis (one-hot) vector, whose $k$-th entry is equal to $1$ and all other entries are zero.

\subsection{Riemannian Gradient and First-Order Optimality Condition for $Q$}
In this subsection, we compute the Riemannian gradient $G_Q$ and update $Q$ by performing a gradient step along the tangent direction followed by a retraction onto the Stiefel manifold~\cite{absil2008optimization}:
\begin{equation}\label{retraction_QR}
\left\{
\begin{aligned}
&[\widetilde Q, R] = \mathrm{qr}\!\left(Q^{(t)} - \eta_t G_{Q^{(t)}}\right),\\
&Q^{(t+1)} = \widetilde Q\,\mathrm{diag}\!\big(\mathrm{sign}(\mathrm{diag}(R))\big),
\end{aligned}
\right.
\end{equation}
where $\eta_t>0$ denotes the step size.
Here, the QR decomposition acts as a retraction operator, mapping the updated iterate back onto the Stiefel manifold, while the diagonal sign correction ensures the uniqueness of the factorization and preserves continuity of the iterates.
The Riemannian update on $\mathrm{St}(d+s,D)$
\[
\begin{aligned}
G_Q =& \nabla_Q F - Q^\mathrm{sym}\big(Q^{T}\nabla_Q F\big) \\
=&  Q\frac{Q^T\nabla_Q F-\nabla_Q^T F Q}{2}+Q^\perp (Q^\perp)^T)  \nabla_Q F,
\end{aligned}
\]
where the Euclidean gradient $\nabla_Q F$ with a specific form by 
{\small 
\[
\begin{aligned}
\nabla_Q F(c,Q,\Theta,\Phi)=&\sum_{i=1}^n \nabla_Q \ell(f(\tau_i)-x_i) \\
= &\sum_{i=1}^n \nabla_y \ell(y)|_{y=f(\tau_i)-x_i}\left[
\tau_i^T,\mathrm{vech}^T(\tau_i\tau_i^T)\Theta
\right].
\end{aligned}
\]}
\begin{proposition}
\label{prop:riemannian_stationary_Q}
If the Riemannian gradient with respect to \(Q\) vanishes, i.e., \(G_Q=\mathbf 0\), then \(Q\) satisfies the first-order optimality condition
\begin{equation}\label{first-order-Q}
\nabla_Q F(c,Q,\Theta,\Phi) - 2Q\Lambda = \mathbf 0,
\qquad
Q^\top Q = I_{d+s},
\end{equation}
where \(\Lambda\in\mathbb{R}^{(d+s)\times(d+s)}\) is a symmetric matrix of Lagrange multipliers.
\end{proposition}

The proof is placed in the appendix. It is worth noting that although the loss function is an affine mapping composed with a convex outer function, the resulting objective is not convex with respect to $Q$ due to the nonconvex geometry induced by the Stiefel manifold constraint.

\subsection{Gradient for $\Theta$ and $c$}

 Since the loss function involves linear transformations with respect to both $c$ and $\Theta$, followed by an outer norm operation, the optimization properties of $c$ and $\Theta$ are inherently coupled. We therefore analyze these two variables jointly in this subsection.

\paragraph{Gradient with respect to $\Theta$}
For each $i$, the mapping $\Theta \;\mapsto\; V\Theta^T \mathrm{vech}(\tau_i\tau_i^T)$ is affine in $\Theta$. Since the loss function $\ell(\cdot)$ is convex, the composition
$\ell\!\left(f_{\Theta,c}(\tau_i)-x_i\right)$
is convex with respect to $\Theta$.
Applying the chain rule, we obtain
\[
\begin{aligned}
&\nabla_\Theta
\sum_{i=1}^n
\ell\!\left(f_{\Theta,c}(\tau_i)-x_i\right)\\
=&
\sum_{i=1}^n
\mathrm{vech}(\tau_i\tau_i^T)
\Big(
 \nabla_y \ell(y-x_i)^T
\Big)
\Big|_{\,y=f_{\Theta,c}(\tau_i)}  V .
\end{aligned}
\]

\paragraph{Gradient with respect to $c$}
Similarly, the inner function is affine in $c$, and the gradient is given by
\[
\nabla_c
\sum_{i=1}^n
\ell\!\left(f_{\Theta,c}(\tau_i)-x_i\right)
=
\sum_{i=1}^n
\nabla_y \ell(y-x_i)
\Big|_{\,y=f_{\Theta,c}(\tau_i)} .
\]

\paragraph{KKT condition}
In this section, we give the first-order optimal condition for our objective, which is a complex nonlinear equation system.
{\small
\begin{equation}\label{first-order-condition}
\left\{
\begin{aligned}
    &P_{T_Q}(\nabla_Q F) = Q (\frac{Q^T\nabla_Q F -\nabla_Q^T F Q}{2} ) + Q^\perp {Q^\perp}^T \nabla_Q F  = {\bf 0},\\
& \nabla_{\Theta} F = \sum_{i=1}^n
\mathrm{vech}(\tau_i\tau_i^T)
\Big(
 \nabla_y \ell(y-x_i)^T
\Big)
\Big|_{\,y=f_{\Theta,c}(\tau_i)}  V,\\
    &\nabla_c F =\sum_{i=1}^n
\nabla_y \ell(y-x_i)
\Big|_{\,y=f_{\Theta,c}(\tau_i)} = {\bf 0},\\
     &\nabla_{\tau_i}F = (U +V\Theta^T (M_{\tau_i}+N_{\tau_i}))^T\nabla_y \ell(y-x_k)|_{y = f(\tau_i)}={\bf 0}, 
\end{aligned}
\right.
\end{equation}}
where $P_{T_Q}(\cdot)$ denotes the orthogonal projection onto the tangent space $T_Q\mathrm{St}(d+s,D)$ of the Stiefel manifold, ensuring that the stationarity condition with respect to $Q$ is compatible with the orthogonality constraint.

Owing to the strong nonlinearity of the above system and the coupling among variables, directly solving the first-order conditions in \eqref{first-order-condition} is intractable in practice. Instead, we adopt a gradient-based optimization strategy, where each variable is updated iteratively using (Riemannian) gradient descent, as detailed in Section~\ref{Alg}.

\section{Convexity Analysis}\label{sec:convexity}
We analyze the convexity of each subproblem with respect to the variables $\Theta, c$, and $\{\tau_k\}$.
When all other variables are fixed, the subproblem in $c$ is convex, since the outer loss function $\ell(\cdot)$ is convex and the inner mapping is affine in c. The same argument applies to the subproblem in $\Theta$.
Consequently, the joint problem
$\min_{\Theta,\,c}\ \sum_{i=1}^n \ell\!\left(f_{\Theta,c}(\tau_i)-x_i\right)$
is a convex optimization problem, as it consists of a sum of convex functions composed with affine mappings. As a result, for fixed $\{\tau_i\}$, any stationary point with respect to $(\Theta,c)$ is globally optimal.


In contrast, the subproblem with respect to $\tau_k$ is more challenging, as the inner transformation $f(\tau_k)$ is nonlinear. Nevertheless, we can still investigate the condition under which the Hessian is positive definite.
Since the overall objective is separable across samples,
\[
F(c,Q,\Theta,\Phi) = \sum_{k=1}^n \ell \big(f(\tau_k)-x_k\big),
\]
the optimization with respect to $\{\tau_k\}_{i=1}^n$ decomposes into independent single-sample subproblems. Consequently, it suffices to analyze the convexity properties of the function $\tau \mapsto \ell \big(f(\tau)-x\big)$ for a single data point.

\paragraph{Convexity of the Projection Problem}
Here, we investigate the optimization problem with respect to the lower-dimensional representation $\{\tau_k\}_{k=1}^n$. Since each subproblem with respect to $\tau_k$ is independent, for notational convenience, we leave out the index $k$ and study on general problem as
\[
\min_{\tau} \ell(f(\tau)-x).
\]
We next derive the second-order derivative $H(\tau)$  with respect to the latent variable $\tau$. Denote $y:=f(\tau)-x$. The Hessian admits the decomposition as
\begin{equation}\label{eq:hess_tau}
\begin{aligned}
H(\tau) = \nabla_{\tau} f(\tau)^{\top}\nabla_{y}^2 \ell(y)\nabla_{\tau} f(\tau)+\nabla_{\tau}^2 f(\tau)
\times_1
\nabla_y \ell(y) ,
\end{aligned}
\end{equation}
Here, $\nabla_{y}^{2}\ell(y)$ denotes the Hessian of the loss with respect to $y$, and $\times_{1}$ represents the mode--$1$ tensor--vector contraction, i.e., the contraction of the first mode of the third-order tensor $\nabla_{\tau}^{2} f(\tau)$ with the vector $\nabla_{y}\ell(y)$ evaluated at $y=f(\tau)-x$. For clarity, we summarize the dimensions of the involved quantities:
$\nabla_{\tau} f(\tau)\in\mathbb{R}^{D\times d}, \nabla_{\tau}^{2} f(\tau)\in\mathbb{R}^{D\times d\times d}, \nabla_{y}\ell(y)\big|_{y=f(\tau)-x}\in\mathbb{R}^{D}.$

The first term in~\eqref{eq:hess_tau} characterizes the curvature contribution
induced by the loss function through the Jacobian of $f$, whereas the second
term captures the intrinsic curvature of the model manifold arising from the
nonlinear mapping $f$. Owing to the convexity of the loss function $\ell$, the
Hessian $\nabla_y^2 \ell(y)\big|_{y=f(\tau)-x}$ is positive semidefinite, which
implies that the first term in~\eqref{eq:hess_tau} is also positive semidefinite.
In particular, for the $\ell_p^p$ loss, the Hessian with respect to $y$ admits
the diagonal form
\[
\nabla_y^2 \ell(y)\big|_{y=f(\tau)-x}
=
p(p-1)\,
{\rm diag}\!\left(
\bigl\{|f(\tau)_i - x_i|^{p-2}\bigr\}_{i=1}^D
\right),
\]
which is positive semidefinite for all $p \ge 1$.

As we restrict $f$ from the quadratic form, the second-order derivative with respect to $\tau$ is constant and given by
\[
\big\{\nabla_{\tau}^2 f(\tau)\big\}_{ruv}
=
2\sum_{s} V_{rs}\,\mathcal{A}_{suv}.
\]
where $\nabla_{\tau}^2 f(\tau) \in {\mathbb R}^{D\times d\times d}$. Therefore, 
\begin{equation}\label{hessian2}
\begin{aligned}
&\nabla_{\tau}^2 f(\tau)
\times_1
\nabla_y \ell(y)\big|_{y=f(\tau)-x}\\
 =& 2\sum_{r=1}^D \{\sum_{t=1}^s V_{rt}\,\mathcal{A}_{tuv}\}  \{\nabla_y \ell(y)\big|_{y=f(\tau)-x} \}_r.
\end{aligned}
\end{equation}

Owing to the presence of the curvature tensor $\mathcal{A}$ and its contraction
with the gradient $\nabla_y \ell(y)$, the definiteness of the term
$
\nabla_{\tau}^2 f(\tau)
\times_1
\nabla_y \ell(y)\big|_{y=f(\tau)-x}
$
cannot be determined directly. Nevertheless, its magnitude can be controlled by
analyzing the interaction between the curvature tensor $\mathcal{A}$ and the
derivative of the loss function. In particular, for the $\ell_p^p$ loss with $p>1$, either a sufficiently small
gradient norm $\nabla_y \ell(y)|_{y=f(\tau)-x}$ (small noise setting) or a small curvature scale of
the tensor $\mathcal{A}$ effectively suppresses the contribution of
\eqref{hessian2}. As a consequence, the aggregated Hessian $H(\tau)$ remains
positive definite over a relatively large region of the parameter space. 

Motivated by this observation, we next establish a theorem that quantitatively
characterizes the region in which $H(\tau)$ is positive definite by analyzing the
structure of the Hessian matrix.



\begin{theorem}[Local convexity radius for $\ell_p^p$-SCQM]\label{thm:conv_radius_lp}
Let $1<p\leq2$ and consider the objective
\[
\min_\tau
 \ell(\tau) =\|f(\tau)-x\|_p^p.
\]
Assume there exist constants $\sigma_0,\rho,A_0>0$ such that, the following conditions hold:
\begin{enumerate}
\item the Jacobian satisfies
$\sigma_{\min}(\nabla_{\tau} f(\tau))\ge \sigma_0$;
\item the residual is bounded away from zero, i.e.,
$\min_j |(f(\tau)-x)_j|^{p-2}\ge \rho$;
\item the quadratic tensor satisfies
$\|\mathcal A\| := \max_s \|\mathcal A_s\| \le A_0$.
\end{enumerate}
Then the Hessian of $\ell(\tau)$ is positive semidefinite throughout the $p-1$ norm ball $\mathcal{N}_{p-1}(x,r_p)$:
\[
\mathcal{N}_{p-1}(x,r_p)
\;:=\;
\left\{
\tau \in \mathcal{T}
\;\middle|\;
\|f(\tau)-x\|_{p-1}
\le r_p
\right\},
\]
where $
r_p
:=
\left(
\frac{(p-1)\rho\sigma_0^2}{2A_0}
\right)^{\!\frac{1}{p-1}} .
$

\end{theorem}

The proof for Theorem~\ref{thm:conv_radius_lp} is placed in the Appendix.
It is worth noting that the assumption $\sigma_{\min}(\nabla_\tau f(\tau)) \ge \sigma_0$ is natural in our setting. Indeed, since $\nabla_\tau f(\tau)=U + V\Theta^{\top}(M_{\tau}+N_{\tau}),$
where $U$ and $V$ have orthonormal columns and satisfy $U^\top V = 0$, we have for any $z $,
\[
\begin{aligned}
\|\nabla_\tau f(\tau) z\|_2^2 = &\|Uz\|_2^2 + \|V\Theta^{\top}(M_{\tau}+N_{\tau})z\|_2^2\\
= &\|z\|_2^2 + \|\Theta^{\top}(M_{\tau}+N_{\tau})z\|_2^2 \ge \|z\|_2^2.
\end{aligned}
\]
This immediately implies $\sigma_{\min}(\nabla_\tau f(\tau)) \ge 1,$ and therefore one may take $\sigma_0 = 1$ without loss of generality. The assumption $\min_j |(f(\tau)-x)_j|^{p-2} \ge \rho$ is mild in practice when $p$ is close to $2$, as long as $(f(\tau)-x)_j\neq 0, \forall j$ and $\rho \in (0,1)$ is chosen sufficiently small. In particular, in the special case $p=2$, the Hessian of the loss reduces to the identity matrix and the above condition holds trivially. Moreover, the boundedness assumption on the quadratic tensor, $\|\mathcal A\| := \max_s \|\mathcal A_s\| \le A_0,$ is natural, since $A_0$ can be taken as the maximum operator norm over all matrix slices $\{\mathcal A_s\}$ of the tensor $\mathcal A$.

\section{Sensitivity analysis}\label{sec:sensitivity}
Due to the high dimensionality of the second-order parameters and the presence of orthogonality constraints, directly conducting a full sensitivity analysis for all variables based on the KKT conditions in Eqn.~\eqref{first-order-condition} is analytically intractable. Instead, we adopt a simplified yet insightful approach by studying the sensitivity of the Fr\'echet mean~\cite{bhattacharya2012nonparametric} with respect to the input samples~$\{x_i\}_{i=1}^n$. This setting can be regarded as a reduced version of our quadratic model, in which linear, curvature terms and manifold constraints are omitted, while the influence of the loss function remains explicit.

The Fr\'echet mean under a general loss function $\ell(\cdot)$ is defined in Eqn.~\eqref{c_optimal}. We view the optimizer $c^*$ as an implicit function of the data $\{x_i\}_{i=1}^n$.  By invoking the implicit function theorem~\cite{krantz2013implicit}, we characterize how $c^*$ responds to small perturbations  $\{\Delta x_i\}_{i=1}^n$ in the observations.



\begin{proposition}[Sensitivity of the Optimal Solution]\label{prop:sensitivity_c}
Let $\ell:\mathbb{R}^D \to \mathbb{R}$ be a convex and twice continuously differentiable loss function, and let
\begin{equation}\label{c_optimal}
c^*(X) := \arg\min_{c \in \mathbb{R}^D} \sum_{i=1}^n \ell(x_i - c),
\end{equation}
denote the optimal solution for the data matrix $X = \{x_i\}_{i=1}^n$. Define the residuals
\[
r_i^* := x_i - c^*, \qquad i=1,\dots,n,
\]
and the aggregated Hessian
\[
H := \sum_{i=1}^n \nabla^2 \ell(r_i^*).
\]

Then, the mapping $X \mapsto c^*(X)$ is locally Fr\'echet differentiable. In particular, for any perturbation $\{\Delta x_i\}_{i=1}^n$, the induced first-order variation of $c^*(X)$ is given by
\begin{equation}\label{robust_result}
\Delta c = H^{-1} \sum_{i=1}^n \nabla^2 \ell(r_i^*) \, \Delta x_i.
\end{equation}
\end{proposition}

The proof for Proposition~\ref{prop:sensitivity_c} is left in the appendix. Based on Proposition~\ref{prop:sensitivity_c}, we specialize the analysis to the $\ell_p^p$ loss with $1 < p \leq 2$ and to the $\ell_2$ loss.  This proposition allows us to elucidate why these loss functions yield estimators that are more robust than those  based on the squared $\ell_2^2$ loss.  In particular, we show that the corresponding estimators possess bounded sensitivity to data perturbations, 
which significantly enhances their robustness to outliers.

\subsection{Robustness Interpretation}
Here, we explain why the $\ell_p^p$  loss for $1<p\leq 2$ leads to enhanced robustness, based on the sensitivity result established in Proposition~\eqref{prop:sensitivity_c}.
For the $\ell_p^p$ loss, the Hessian $\nabla^2 \ell(r_i^\ast)$ is diagonal, with the $k$-th diagonal entry given by
\[
\nabla^2 \ell(r_i^\ast) = p(p-1) {\rm diag} ( \{|r_{ik}^\ast|^{p-2} \}_{k=1}^D ).
\]

For $p=2$, the loss reduces to the squared Euclidean loss $\ell(r)=\|r\|_2^2$, and the Hessian is constant: $\nabla^2 \ell(r_i^\ast) = 2 I_D.$ Consequently, the aggregated Hessian satisfies $H = 2n I_D$, and the sensitivity formula in Proposition~\ref{prop:sensitivity_c} simplifies to
$
\Delta c
=
-\frac{1}{n}
\sum_{i=1}^n
\Delta x_i.
$
This shows that the optimal intercept $c^\ast$ responds linearly and uniformly to perturbations in the data, without any attenuation from the residual magnitudes.

For $1 < p \leq 2$, the Hessian weights depend on the residual magnitudes through $|r_{ik}^\ast|^{p-2}$.
In particular, large residuals receive smaller weights, while small residuals are emphasized.
As a result, the perturbation $\Delta x_i$ is effectively reweighted in the sensitivity formula, leading to
\[
\Delta c = \,H^{-1}
\sum_{i=1}^n \mathrm{diag}(  \{|r^*_{ik}|^{p-2}\}_{k=1}^D ) \Delta x_i.
\]
We observe that a perturbation $\Delta x_{ik}$ in the $k$-th coordinate exerts a strong influence on the estimator when the corresponding residual $|r_{ik}^\ast|$ is small, while its influence becomes attenuated when $|r_{ik}^\ast|$ is large. This behavior arises because the sensitivity is weighted by the factor $|r_{ik}^\ast|^{p-2}$, whose exponent satisfies $p-2\leq0$ for $1<p\leq2$. Consequently, large residuals receive diminishing weight in the sensitivity propagation, which explains the enhanced robustness of the $\ell_p^p$ loss compared to the squared Euclidean loss.

Next, we investigate the robustness properties of the $\ell_2$ norm.
For the $\ell_2$ loss $\ell(r)=\|r\|_2$, when the residual $r \neq 0$, the Hessian admits the closed-form expression
\[
\nabla^2 \ell(r)
=
\frac{1}{\|r\|_2}
\left(
I - \frac{r r^{\top}}{\|r\|_2^2}
\right)
=
\frac{1}{\|r\|_2}\, P_{r^\perp},
\]
where $P_{r^\perp}$ denotes the orthogonal projection matrix onto the subspace orthogonal to $r$. Substituting this expression into~\eqref{robust_result} yields the first-order sensitivity of the optimal intercept:
\begin{equation}\label{ell2_delta_c}
\Delta c
=
H^{-1}
\sum_{i=1}^n
\frac{1}{\|r_i^\ast\|_2}
\, P_{{r_i^\ast}^\perp}
\, \Delta x_i .
\end{equation}
This representation reveals two intrinsic robustness mechanisms of the $\ell_2$ loss.
First, perturbations associated with large residuals are attenuated by the factor $1/\|r_i^\ast\|_2$, thereby reducing the influence of outliers.
Second, perturbations in the direction of the residual $r_i^\ast$ do not affect $\Delta c$, since they are annihilated by the projection operator $P_{{r_i^\ast}^\perp}$.
Together, these properties explain the improved stability of $\ell_2$-based estimators relative to the squared Euclidean loss.


\section{Algorithm}\label{Alg}
In this section, we present a gradient descent algorithm for solving the proposed generalized subspace-constrained quadratic model. The algorithm is highly flexible and accommodates a broad class of differentiable (or subdifferentiable) loss functions; adapting it to a specific loss only requires replacing the gradient (or subgradient) term $\nabla_y \ell(y - x_i)$ accordingly.

Before introducing the algorithm, we describe two technical mechanisms that are incorporated to improve its convergence behavior. First, we employ an orthonormal retraction based on an aligned QR decomposition, which preserves the Stiefel manifold constraint and stabilizes the gradient flow. Second, we adopt an asynchronous learning-rate adjustment strategy, which enhances numerical stability and accelerates the optimization process.

\paragraph{Orthonormal retraction via aligned QR}
Due to the orthogonality constraint imposed on $Q$, the Euclidean gradient update reduces to an update along the Riemannian gradient direction $G_Q$, which is obtained by projecting the Euclidean gradient onto the tangent space at $Q$. To maintain the orthonormality constraint, the updated matrix is then retracted back onto the Stiefel manifold via a QR decomposition~\cite{absil2008optimization}. However, the QR decomposition is not unique: for a given matrix, there may exist two orthonormal matrices $Q$ and $Q'$, together with two upper triangular matrices $R$ and $R'$, such that $QR = Q'R'$. To remove this ambiguity, we impose a standard sign convention on the diagonal entries of the upper triangular factor, requiring them to be positive, as
discussed in~\cite{golub2013matrix}. Under this convention, the QR decomposition is unique, and the associated retraction map is continuous. Consequently, the update sequence $\{Q^{(t)}\}$ satisfies
\[
\lim_{\eta_t \to 0} \bigl\| Q^{(t+1)} - Q^{(t)} \bigr\|_{\rm F} = 0,
\]
ensuring that successive iterates vary smoothly as the step size diminishes.
This continuity property is essential for guaranteeing a monotonic decrease of
the objective function during the gradient descent process and for maintaining
the stability of the optimization on the Stiefel manifold.
\begin{algorithm}[h!]
\caption{Riemannian Gradient Descent for SCQM }
\label{alg:RGD_QMF_tight}
\begin{algorithmic}[1]
\REQUIRE
$X=[x_1,\dots,x_n]\in\mathbb{R}^{D\times n}$; 
max iterations $T$; tolerance $\varepsilon$; 
initial step sizes $(\eta_Q,\eta_\Theta,\eta_c,\eta_\tau)$;
loss $\ell$ with (sub)gradient $\partial\ell(\cdot)$;
retraction $\mathrm{Retr}_{\mathrm{St}}(\cdot)$ and $\mathrm{sym}(A)=\tfrac12(A+A^T)$.
\ENSURE
$Q=[U,V]\in\mathrm{St}(d+s,D)$, $\Theta$, $c$, $\{\tau_i\}_{i=1}^n$.

\STATE \textbf{Initialize:}
$c^{(0)}=\tfrac{1}{n}X\mathbf{1}_n$; compute PCA of $X-c^{(0)}\mathbf{1}_n^T$ and set
$Q^{(0)}=[U^{(0)},V^{(0)}]\in\mathrm{St}(d+s,D)$;
$\tau_i^{(0)}=(U^{(0)})^T(x_i-c^{(0)})$;
$\Theta^{(0)}\sim\mathcal{N}(0,\sigma^2 I)$.

\FOR{$t=0,\dots,T-1$}
\STATE \textbf{Residuals and weights:} \\
\textbf{for} $i=1,\dots,n$
\[
\begin{aligned}
&\phi_i^{(t)}=\mathrm{vech} \big(\tau_i^{(t)}\tau_i^{(t)T}\big),\\
&f_i^{(t)}=c^{(t)}+U^{(t)}\tau_i^{(t)}+V^{(t)}\Theta^{(t)T}\phi_i^{(t)},\\
&r_i^{(t)}=f_i^{(t)}-x_i,\qquad g_i^{(t)}\in\partial\ell \big(r_i^{(t)}\big).
\end{aligned}
\]
\STATE \textbf{end for}

\STATE \textbf{Euclidean gradients:}
\[
\begin{aligned}
&\nabla_c F^{(t)}=\sum_{i=1}^n g_i^{(t)}, \\
&\nabla_\Theta F^{(t)}=\sum_{i=1}^n \phi_i^{(t)}\big(V^{(t)T}g_i^{(t)}\big)^T,\\
&\nabla_Q F^{(t)}=\sum_{i=1}^n g_i^{(t)}
\begin{bmatrix}
\tau_i^{(t)T} & \phi_i^{(t)T}\Theta^{(t)}
\end{bmatrix}.
\end{aligned}
\]

\STATE \textbf{Riemannian step on $\mathrm{St}(d+s,D)$:}
\[
\begin{aligned}
&G_Q^{(t)}=\nabla_Q F^{(t)}-Q^{(t)}\mathrm{sym}\!\big(Q^{(t)T}\nabla_Q F^{(t)}\big),\\
&Q^{(t+1)}=\mathrm{Retr}_{\mathrm{St}} \big(Q^{(t)}-\eta_Q^{(t)}G_Q^{(t)}\big),
\end{aligned}
\]
extract $Q^{(t+1)}=[U^{(t+1)},V^{(t+1)}]$.

\STATE \textbf{Euclidean updates:}
\[
\begin{aligned}
&\Theta^{(t+1)}=\Theta^{(t)}-\eta_\Theta^{(t)}\nabla_\Theta F^{(t)},\\
&c^{(t+1)}=c^{(t)}-\eta_c^{(t)}\nabla_c F^{(t)}.
\end{aligned}
\]

\STATE \textbf{Update latent coordinates:} \\
\textbf{for} $i=1,\dots,n$
\[
\begin{aligned}
&J_{\tau_i}^{(t)}=\frac{\partial f_i^{(t)}}{\partial \tau_i}\bigg|_{(U^{(t+1)},V^{(t+1)},\Theta^{(t+1)},\tau_i^{(t)})},\\
&\tau_i^{(t+1)}=\tau_i^{(t)}-\eta_\tau^{(t)}\,J_{\tau_i}^{(t)T} g_i^{(t)}.
\end{aligned}
\]
\STATE \textbf{end for}
\ENDFOR
\end{algorithmic}
\end{algorithm}
\paragraph{Asynchronous learning-rate adjustment}
An appropriate choice of the learning rate is crucial for both convergence behavior and computational efficiency. However, the optimal learning rate is influenced by multiple factors, including the local geometric properties of the objective function at the current iterate and the curvature induced by the constraint manifold. Asynchronous learning-rate adjustment enables the use of distinct learning rates for different parameter blocks—such as $\eta_Q$ for the subspace variable $Q$ and $\eta_\Theta$ for the quadratic parameters $\Theta$—allowing each component to be updated in accordance with its own geometric and optimization characteristics.

To ensure an effective learning rate, we adopt an asynchronous learning-rate adjustment strategy. Specifically, we initialize separate learning rates $\eta_c, \eta_Q, \eta_\Theta,$ and $\eta_\tau$ for the variables $c$, $Q$, $\Theta$, and ${\tau_i}$, respectively. During each update, the learning rate associated with a given variable is adjusted independently based on a sufficient decrease criterion.

We employ a backtracking line search strategy based on the Armijo rule to adaptively determine suitable step sizes during optimization~\cite{nocedal2006numerical}. 
Specifically, when updating the variable $c$, we first take a tentative gradient step and evaluate the objective function $F\!\left(c - \eta_c \nabla_c F,\; \Theta,\; Q,\; \{\tau_i\}\right)$. This value is compared against the sufficient decrease condition
$F(c,\Theta,Q,\{\tau_i\}) - \alpha \eta_c \big\|\nabla_c F(c,\Theta,Q,\{\tau_i\})\big\|_{\mathrm{F}}^2,$
where $\alpha \in (0,1)$ is a prescribed constant.
If the condition is satisfied, the step size $\eta_c$ is accepted; otherwise, it is reduced, for example by setting $\eta_c \leftarrow \eta_c / 2$, and the test is repeated.
The same backtracking procedure is applied independently to the updates of $Q$, $\Theta$, and $\{\tau_i\}$. 
This block-wise adaptive step-size strategy allows each parameter group to adjust to the local geometry of the objective function and the associated constraint manifold, thereby improving numerical stability and accelerating the convergence of the overall optimization algorithm.

\paragraph{Learning process} The learning process can be summarized as: Given a data matrix $X \in \mathbb{R}^{D \times n}$, the algorithm jointly estimates the latent representations $\{\tau_i\}_{i=1}^n$, the quadratic mapping parameters $\Theta$, the mean vector $c$, and an orthonormal basis $Q \in \mathrm{St}(d+s, D)$. 

The algorithm initializes $c$ as the column mean of $X$, and initializes Q using the leading $d+s$ left singular vectors of the centered data matrix $X - c\mathbf{1}_n^T$. Each latent variable $\tau_i$ is initialized by projecting the centered data point $x_i - c$ onto the first $d$ columns of $Q$, while $\Theta$ is initialized randomly. 

At each iteration, the Euclidean gradients with respect to all variables—$\nabla_Q F, \nabla_\Theta F, \nabla_c F$, and $\{\nabla_{\tau_i} F\}_{i=1}^n$—are first computed. To enforce the orthonormality constraint on $Q$, the gradient $\nabla_Q F$ is projected onto the tangent space of the Stiefel manifold, yielding the Riemannian gradient direction. A Riemannian gradient step is then performed, followed by a retraction onto the Stiefel manifold via QR decomposition. A backtracking line search is applied along the retracted Riemannian gradient direction to determine a suitable step size.
For the unconstrained parameters $\Theta$ and $c$, standard Euclidean gradient updates are employed.

For the update of each $\tau_i$, we also employ an asynchronous backtracking line search to determine an acceptable step size that ensures sufficient decrease in the objective value. In other words, the step size associated with each $\tau_i$ is selected independently.
Each latent variable $\tau_i$ is then updated via a gradient step that explicitly incorporates the structured matrices $M_{\tau_i}$ and $N_{\tau_i}$, which arise from the quadratic term of the model. The procedure is iterated until convergence or until a prescribed maximum number of iterations is reached, yielding a structured quadratic approximation that minimizes the chosen loss function.
\section{Numerical Experiments}\label{sec:experiments}

\begin{figure}[t]
    \centering
\includegraphics[width=\linewidth]{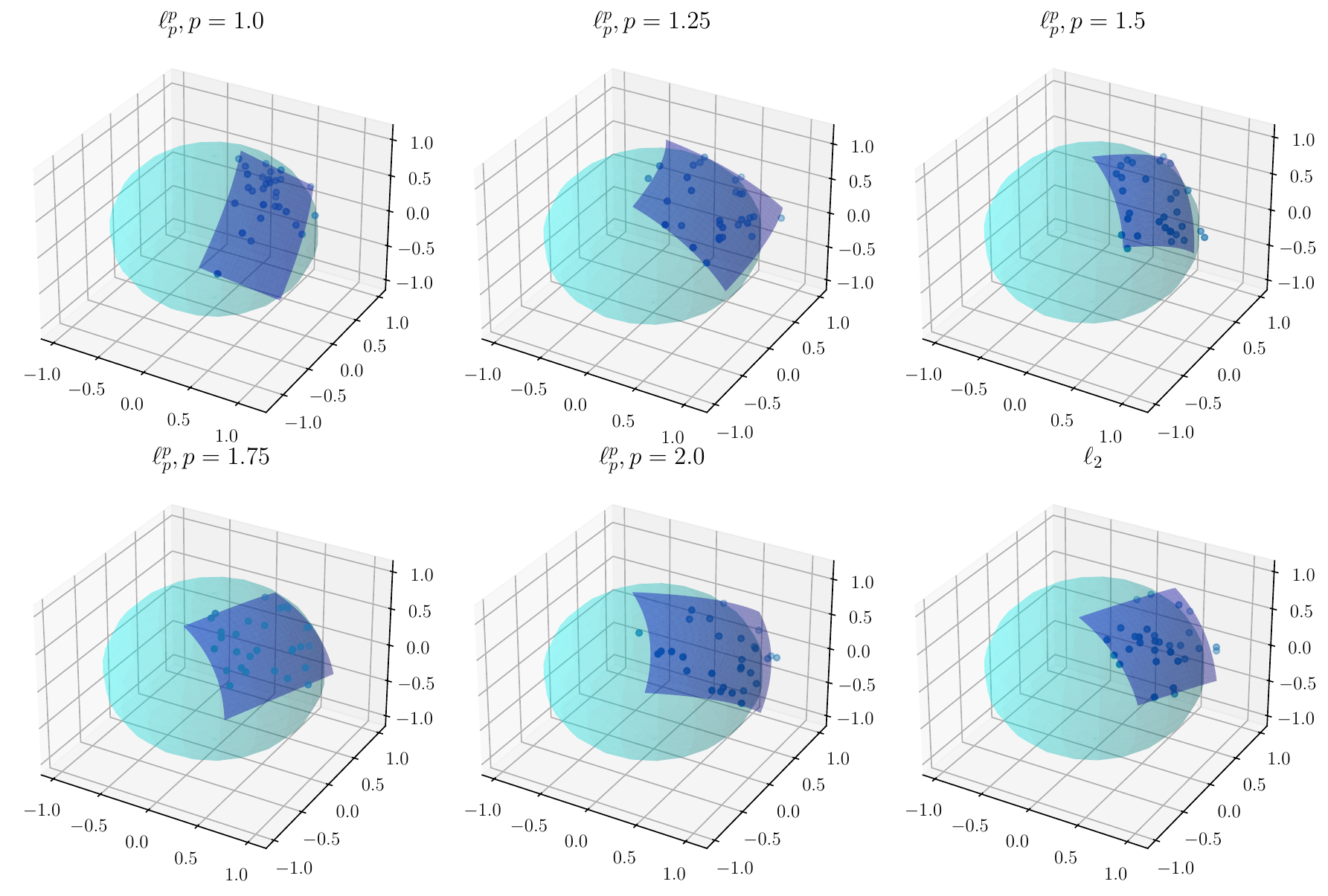}
    \caption{Illustration of the fitted curves and projection points obtained using $\ell_p^p$ losses with different values of $p$.}
\label{fig:robustness2_3d}
\end{figure} 

In this section, we demonstrate how the proposed model can be applied to the task of manifold denoising, or more generally, manifold data refinement. Unlike the toy example in Section~\ref{toy}, where different noise models are paired with their corresponding loss functions, here we evaluate all methods under the same noise distribution. Specifically, we assess the performance of various methods under additive Gaussian noise with different noise scales.

In this experimental setting, we assume that the observed dataset $\{x_i\}$ is generated by corrupting a noise-free dataset $\{\tilde{x}_i\}$ with Gaussian noise $\{\epsilon_i\}$, such that
\begin{equation}\label{noisy_observation}
x_i = \tilde{x}_i + \epsilon_i, \quad \text{where} \quad \tilde{x}_i \sim \mathcal{M}, \ \epsilon_i\sim {\cal N}({\bf 0}, \sigma^2 I).
\end{equation}

We implement data refinement in two steps. First, for each observation \( x_i \), we identify its \( K \) nearest neighbors, denoted as \( \mathcal{N}_{x_i} \). Second, we learn a subspace-constrained quadratic matrix factorization model to find a local representation \( \widehat{f} \), which is parameterized by \( \widehat{Q}, \widehat{\Theta}, \widehat{c} \).
Next, we project \( x_i \) back onto the learned subspace \( \widehat{f} \) by solving the following optimization problem:
\[
P(x_i) = \arg\min_{x \in \{ \widehat{f}(\tau),\tau\in {\mathbb R}^d\}} \|x_i - x\|_p^p.
\]
Finally, we evaluate the performance of the manifold approximation by measuring the empirical mean squared error:
\[
\mathcal{E} = \frac{1}{N} \sum_{i=1}^N \|\tilde{x}_i - P(x_i)\|_2^2.
\]
A smaller value of \( \mathcal{E} \) indicates a stronger recovery capability of the manifold approximation method.


\subsection{Simulation with Spherical Data in ${\mathbb R}^3$}

\begin{table*}[t]
\centering
\caption{Mean squared error (MSE) of different models under varying noise levels.\label{MSE_compare}}
\resizebox{2\columnwidth}{!}{
\begin{tabular}{c|cccccc|cccccc|ccc}
\toprule
& \multicolumn{6}{c|}{Quadratic model: $\Theta\neq \bf 0$} 
& \multicolumn{6}{c|}{Linear model: $\Theta = \bf 0$} 
& \multicolumn{3}{c}{Other methods} \\
\cmidrule(lr){2-7} \cmidrule(lr){8-13} \cmidrule(lr){14-16}
& \multicolumn{5}{c|}{$\ell_p^p$} & $\ell_2$ 
& \multicolumn{5}{c|}{$\ell_p^p$} & $\ell_2$ 
& SPH & MFIT & MLS \\ \hline
\diagbox{$\sigma$}{$p$}& $1.00$ & $1.25$ & $1.50$ & $1.75$ & $2.00$ & $2.00$
& $1.00$ & $1.25$ & $1.50$ & $1.75$ & $2.00$ & $2.00$
&--  &--  & -- \\
\midrule
$0.03$ & 0.009 & 0.002 & 0.002 & 0.001 & 0.001 & 0.002 & 0.016 & 0.016 & 0.014 & 0.013 & 0.013 & 0.013 & 0.002 & 0.018 & 0.001 \\
$0.06$ & 0.014 & 0.006 & 0.006 & 0.005 & 0.005 & 0.006 & 0.021 & 0.020 & 0.017 & 0.015 & 0.015 & 0.014 & 0.018 & 0.022 & 0.005 \\
$0.09$ & 0.020 & 0.016 & 0.017 & 0.016 & 0.016 & 0.015 & 0.023 & 0.022 & 0.021 & 0.021 & 0.021 & 0.021 & 0.044 & 0.038 & 0.014 \\
$0.12$ & 0.030 & 0.023 & 0.025 & 0.022 & 0.021 & 0.020 & 0.033 & 0.035 & 0.027 & 0.025 & 0.024 & 0.026 & 0.084 & 0.028 & 0.016 \\
$0.15$ & 0.070 & 0.061 & 0.064 & 0.067 & 0.071 & 0.053 & 0.072 & 0.074 & 0.061 & 0.056 & 0.055 & 0.057 & 0.180 & 0.061 & 0.053 \\
$0.18$ & 0.083 & 0.100 & 0.144 & 0.132 & 0.137 & 0.099 & 0.087 & 0.085 & 0.082 & 0.084 & 0.087 & 0.090 & 0.262 & 0.113 & 0.085 \\
$0.21$ & 0.126 & 0.139 & 0.232 & 0.216 & 0.204 & 0.140 & 0.129 & 0.140 & 0.133 & 0.125 & 0.126 & 0.132 & 0.317 & 0.140 & 0.122 \\
\bottomrule
\end{tabular}
}
\end{table*}

\begin{figure}[t]
    \centering
\includegraphics[width=\linewidth]{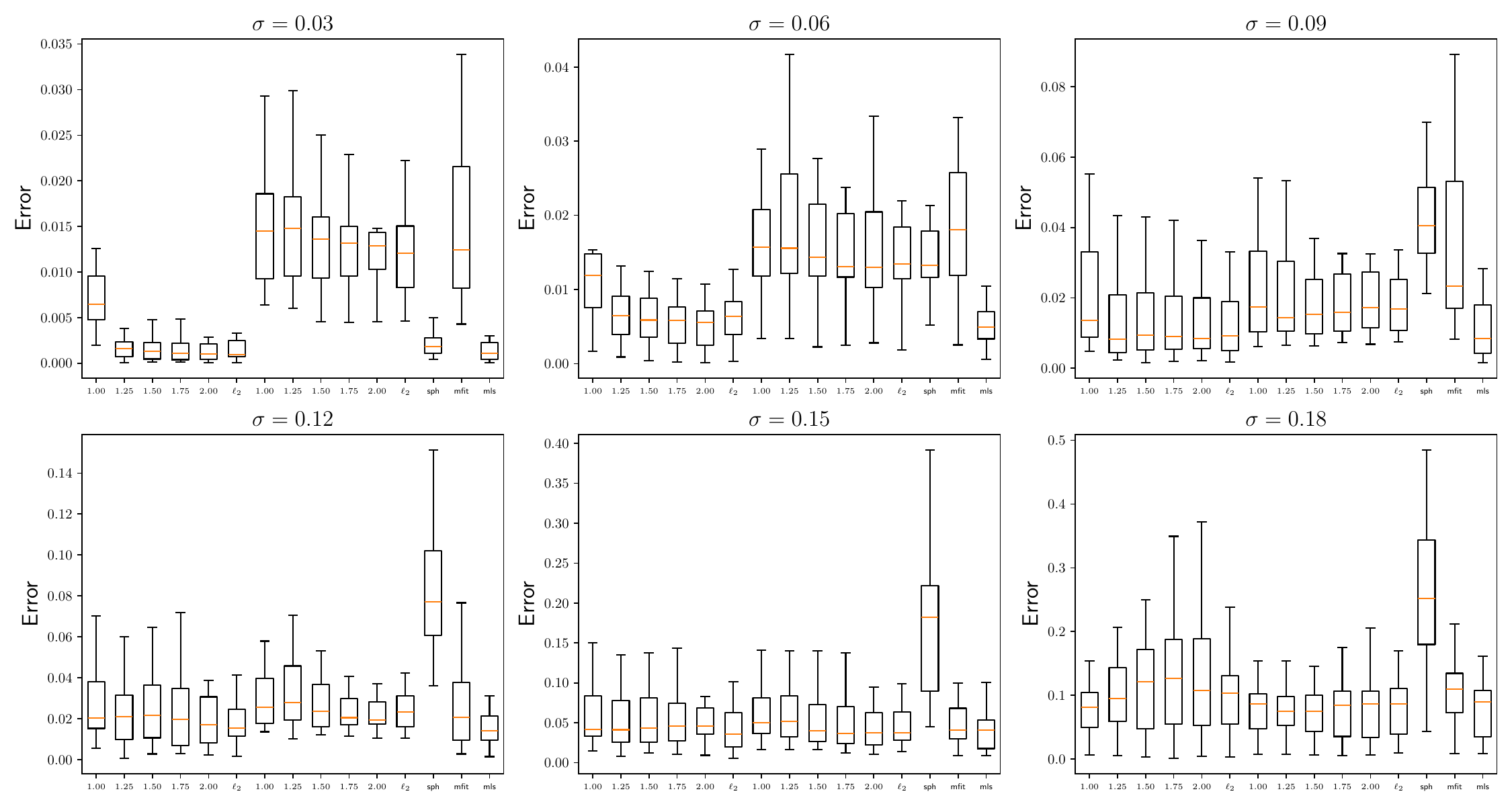}
    \caption{Performance is compared across models and noise levels using boxplots of the squared reconstruction error, $\|\tilde{x}_i - P(x_i)\|_2^2$, computed over all samples.}
    \label{boxplots}
\end{figure}

Here, we compare the performance of the SCQM model with that of its competitors, including SPH, MFIT, and MLS, under different loss functions and across varying noise levels. We uniformly sample 300 points on the 3D sphere as the ground-truth signals ${\tilde{x}_i}$. Noisy observations are then generated according to~\eqref{noisy_observation}. We evaluate model performance under multiple noise settings by varying the noise standard deviation $\sigma \in \{0.05, 0.10, 0.15, 0.20\}$. For each noise level, we test $\ell_p^p$ loss functions with $p \in \{1.0, 1.25, 1.50, 1.75, 2.0\}$, as well as the $\ell_2$ loss. For fairness and completeness, all methods are implemented within a local neighborhood of each sample $x_i$, where the neighborhood contains exactly $30$ neighboring samples.
The per-sample squared error $\|\tilde{x}_i - P(x_i)\|_2^2$ is summarized using boxplots for each combination of model and noise level, as shown in Fig.~\ref{boxplots} and Fig.~\ref{fig:robustness2_3d}.

From the observations in Table~\ref{MSE_compare} and Figure~\ref{boxplots}, we can conclude that :
\begin{itemize}
\item When an appropriate value of 
$p$ is selected, our method consistently outperforms state-of-the-art approaches, including SPH, MFIT, and MLS, across a wide range of noise levels.

\item In the low-noise regime (e.g., $\sigma < 0.15$), the quadratic model consistently outperforms the corresponding linear model, which in turn validates the effectiveness of incorporating quadratic terms. However, as the noise level increases, the performance of the quadratic model deteriorates. For instance, at $\sigma = 0.18$, the quadratic model underperforms the linear model, likely due to overfitting, as it captures spurious patterns arising from the noise rather than the underlying signal.

\item For the quadratic model under the $\ell_p^p$ loss, larger values of $p$ are favorable in the low-noise regime (e.g., $\sigma < 0.12$). However, as the noise level further increases, smaller values of $p$ become more advantageous. This observation highlights a fundamental trade-off between accurate signal projection and robustness to noise. Although minimizing the $\ell_p^p$ loss with $p = 1$ does not recover the true signal under the minimum-distance projection criterion, it exhibits the strongest robustness among the considered methods.

\item The quadratic model with the $\ell_2$ loss exhibits stable and well-balanced performance across different noise levels, demonstrating its ability to effectively handle noise and recover the true projection in Euclidean space, as it avoids the additional sensitivity introduced by higher-order power operations.

\item When the noise level is small, all methods perform well. However, the $\ell_p^p$ loss with $p=1$ consistently underperforms relative to the other metrics, particularly at $\sigma=0.05$ and $\sigma=0.10$. This is likely due to a mismatch between the noise distribution and the $\ell_1$ metric; in other words, an $\ell_1$-based projection does not coincide with the true projection under the Euclidean distance. As the noise level increases, the $\ell_2$ model and the $\ell_p^p$ models with smaller values of $p$ tend to outperform the alternatives. This behavior can be attributed to improved robustness: these losses are less sensitive to large-variance noise and outlier-contaminated samples, leading to more stable performance under heavier noise.

\end{itemize}

\begin{figure}[t]
\centering
\includegraphics[width=0.85\linewidth]{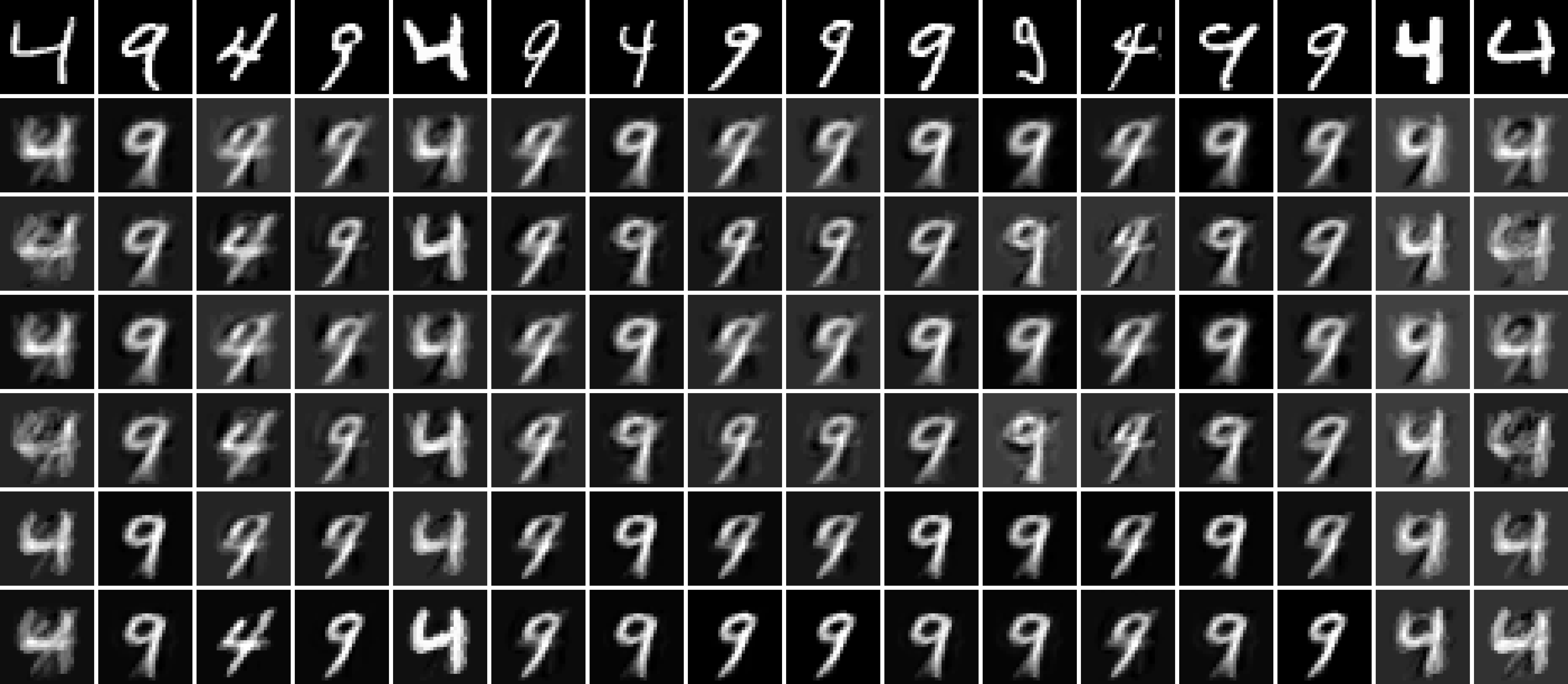}
\caption{Comparison of reconstruction methods under different loss functions and model settings. Each row displays 16 reconstructed images. From top to bottom, the rows correspond to the original images, linear model with $\ell_2$ loss, SCQM with $\ell_2$ loss, linear model with $\ell_p^p$ loss ($p=2$), SCQM with $\ell_p^p$ loss ($p=2$), linear model with $\ell_p^p$ loss ($p=1$), and SCQM with $\ell_p^p$ loss ($p=1$), respectively. }
\label{fig:sqmf_comparison}
\end{figure}

\subsection{Performance on Real World Dataset}
Here, we demonstrate the performance of the robust SCQM on a real-world dataset using handwritten MNIST~\cite{lecun2002gradient} images. We randomly select 100 samples consisting of two easily confused digits, namely `4' and `9'. The original images are projected onto a low-dimensional manifold with latent dimension $d=2$. We visualize the reconstructed images corresponding to these projections and examine the reconstruction details across different model settings.

We evaluate three SCQM models with different loss functions, namely $\ell_2$ (third row), $\ell_2^2$ (fifth row), and $\ell_1$ (seventh row), as shown in Fig.~\ref{fig:sqmf_comparison}. In addition, we evaluate the corresponding linear variants by setting $\Theta=\mathbf{0}$ throughout the learning process, thereby removing the quadratic component. These linear counterparts are displayed in the second, fourth, and sixth rows, corresponding to the three loss functions, respectively. This comparison serves as an ablation study to systematically assess the effectiveness and contribution of the quadratic term within the SCQM framework.
\begin{figure}[t]
\centering
\includegraphics[width=0.95\linewidth]{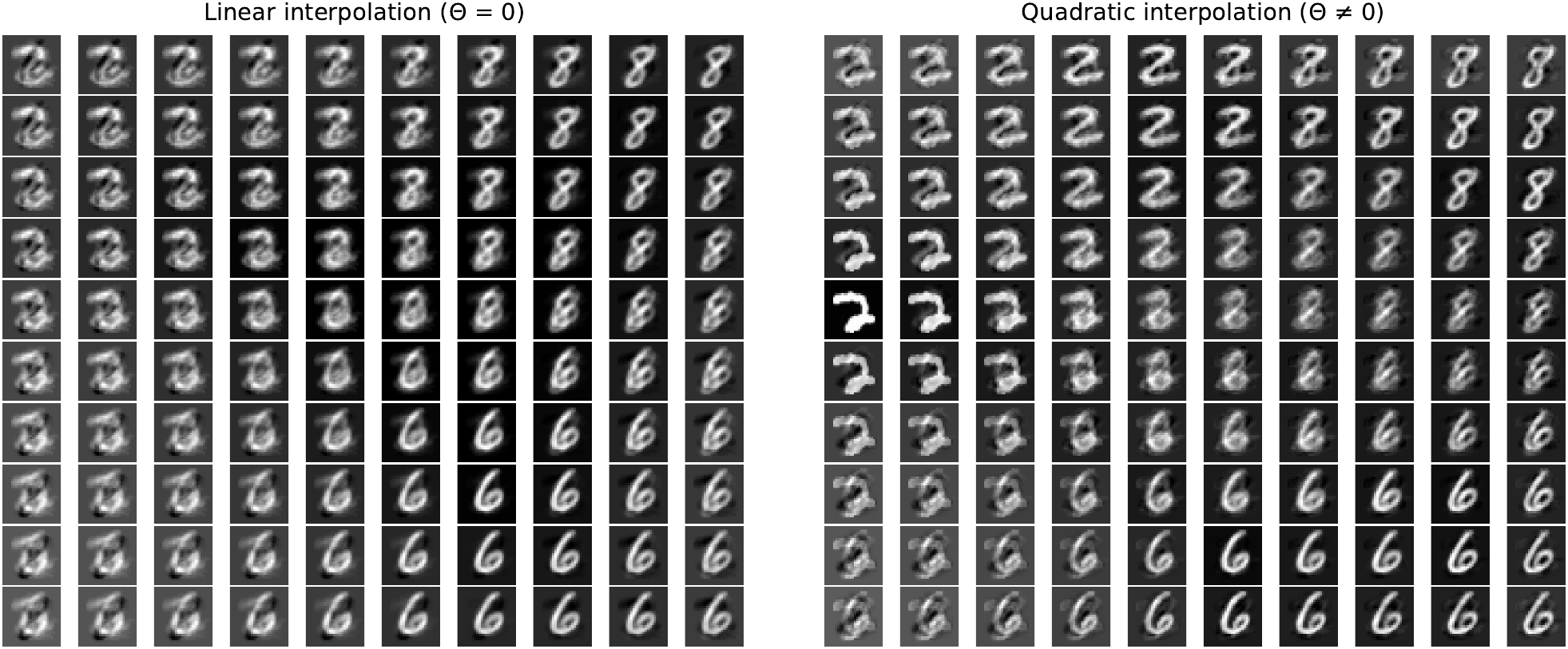}
\caption{Visualization of latent-space ($d=2$) interpolation  for the linear model ($\Theta = \bf 0$) and the quadratic model ($\Theta \neq \bf 0$), learned from data consisting of the three digits `2', `6', and `8'.  }
\label{fig:interpolation}
\end{figure}

From Fig.~\ref{fig:sqmf_comparison}, we draw the following conclusions.
First, by comparing the linear and quadratic variants—specifically, the second versus third rows, the fourth versus fifth rows, and the sixth versus seventh rows—we observe that the inclusion of the quadratic term significantly improves the discrimination between the digits `4' and `9' with the improvement being particularly evident in the second column from the right.
Second, the $\ell_1$ loss and the $\ell_2$ loss consistently outperform the $\ell_2^2$ loss (SQMF discussed in~\cite{zhai2025subspace}), producing sharper and more visually coherent reconstructions. This highlights their superior robustness and reconstruction capability in the presence of ambiguous or overlapping digit structures.

\subsection{Interpolation via SCQM}
In addition, we demonstrate that SCQM can also be used as an interpolation model for generating new high-dimensional data. Since we have learned a mapping $\widehat{f} : \mathbb{R}^d \rightarrow \mathbb{R}^D$, uniformly sampling points in the latent space $\mathbb{R}^d$ allows us to visualize the global landscape of $\widehat{f}$ through its outputs in the ambient space.
This interpolation procedure further serves as a means to evaluate the expressive capability of different models, as it reveals how well each learned mapping captures the underlying manifold structure.

To validate the strong expressive capability of the proposed SCQM in comparison with its linear counterpart in the latent space. We randomly select images corresponding to the digits `2', `6', and `8', and learn an SCQM model $\widehat{f}$ with latent dimension $d=2$ and quadratic dimension $s=20$. We then perform interpolation in the latent space by uniformly sampling between the minimum and maximum values along each latent dimension, resulting in a grid of latent points $\{\tau_k\}$. Each interpolation point is mapped back to the image space via the learned decoder $\widehat{f}:\tau \mapsto I$, and the reconstructed images are visualized at their corresponding latent locations. The results are shown in Fig.~\ref{fig:interpolation}.
As illustrated, the quadratic component plays a crucial role in modeling nonlinear variations of the data manifold. While the linear model produces linear transitions, the quadratic model yields smoother and more continuous interpolations, better capturing the intrinsic geometric structure of the data.

\section{Conclusion}\label{sec:conclusion}
This work investigates a robust quadratic fitting framework that generalizes subspace-constrained quadratic factorization. This generalization provides a promising approach for addressing scenarios with a limited number of observed data points and for relaxing the local flatness assumption inherent in linear subspace models. We propose a gradient descent method to solve the resulting robust quadratic fitting problem and conduct a sensitivity analysis for the Fréchet mean problem, which can be viewed as a special case within the quadratic function class. Numerical experiments are conducted to validate the effectiveness of the proposed framework.

Several directions remain open for future investigation.
First, the statistical properties of the proposed quadratic model have not been analyzed. Future work could establish non-asymptotic results~\cite{wainwright2019high} with respect to the sample size $n$, intrinsic dimension $d$, and additional parameters such as the loss exponent $p$.
Second, although the gradient descent algorithm exhibits stable behavior in our experiments, its theoretical convergence properties have not been studied. Providing convergence guarantees under appropriate conditions is an important direction for future research.
Third, our current analysis focuses on assessing the benefit of incorporating a curvature (quadratic) term relative to the linear case. Extending the framework to more expressive models and exploring richer nonlinear structures may further improve performance and is another promising avenue for future work.


\bibliographystyle{unsrt}
\bibliography{sample}
\end{document}